\documentclass{article}

\usepackage[preprint]{tmlr_arxiv}

\usepackage[utf8]{inputenc} 
\usepackage[T1]{fontenc}    
\usepackage{url}            
\usepackage{booktabs}       
\usepackage{amsfonts}       
\usepackage{nicefrac}       
\usepackage{microtype}      
\usepackage{xcolor}         
\usepackage{graphicx}
\usepackage{subcaption}
\usepackage{amsmath}
\usepackage{amsthm}
\usepackage{multirow}
\usepackage{amssymb}
\usepackage{wrapfig} 
\usepackage{subfiles}
\usepackage[linewidth=1pt]{mdframed}
\usepackage[]{authblk}

\usepackage[acronym,nohypertypes={acronym,notation}]{glossaries}
\glsdisablehyper
\newacronym{dnn}{DNN}{Deep Neural Network}
\newacronym{cnn}{CNN}{Convolutional Neural Network}
\newacronym{nn}{NN}{Neural Network}
\newacronym{mi}{MI}{Mutual Information}
\newacronym{ib}{MI}{Information Bottleneck}
\newacronym{icb}{ICB}{Input Compression Bound}
\newacronym{nas}{NAS}{Neural Architecture Search}
\newacronym{cpc}{CPC}{Contrastive Predictive Coding}
\newacronym{ntk}{NTK}{Neural Tangent Kernel}
\newacronym{awgn}{AWGN}{Additive White Gaussian Noise}
\newacronym{fgsm}{FGSM}{Fast Gradient-Sign Method}
\newacronym{mse}{MSE}{Mean-Squared Error}
\newacronym{loo}{LOO}{Leave-One-Out}
\newacronym{mdl}{MDL}{Minimum Description Length}
\newacronym{gd}{GD}{Gradient Descent}
\newacronym{ge}{GE}{Generalization Error}

\usepackage{amsmath,amsfonts,bm}

\def\eqref#1{equation~\ref{#1}}

\def\1{\bm{1}}

\DeclareMathAlphabet{\mathsfit}{\encodingdefault}{\sfdefault}{m}{sl}
\SetMathAlphabet{\mathsfit}{bold}{\encodingdefault}{\sfdefault}{bx}{n}

\newcommand{\E}{\mathbb{E}}

\theoremstyle{definition}

\title{Bounding generalization error with input compression: \\
An empirical study with infinite-width networks}

\author[1,3]{Angus Galloway}
\author[4,5]{Anna Golubeva}
\author[1,3]{Mahmoud Salem}
\author[2,3]{Mihai Nica}
\author[6]{Yani Ioannou}
\author[1,2,7]{Graham W.~Taylor}
\affil[1]{School of Engineering, University of Guelph, ON, Canada}
\affil[2]{Department of Mathematics and Statistics, University of Guelph, ON, Canada}
\affil[3]{Vector Institute for Artificial Intelligence, ON, Canada}
\affil[4]{The NSF AI Institute for Artificial Intelligence and Fundamental Interactions}
\affil[5]{Department of Physics, Massachusetts Institute of Technology, MA, USA}
\affil[6]{Schulich School of Engineering, University of Calgary, AB, Canada}
\affil[7]{Canada CIFAR AI Chair}
\affil[ ]{Correspondence to \textit{angusdigalloway@gmail.com, \{gallowaa, gwtaylor\}@uoguelph.ca, golubeva@mit.edu}}

\def\biblio{\bibliographystyle{abbrvnat}\bibliography{tmlr_bibliography}}

\begin{document}
\def\biblio{}

\maketitle

\begin{abstract}

Estimating the~\gls{ge} of \glspl{dnn} is an 
important task that often relies on availability of held-out data. 
The ability to better predict~\gls{ge} based 
on a single training set may yield overarching \gls{dnn} design 
principles to reduce a reliance on trial-and-error, 
along with other performance assessment advantages. 
In search of a quantity relevant to~\gls{ge}, we 
investigate the~\gls{mi} between the input and final layer 
representations, using the infinite-width~\gls{dnn} limit 
to bound~\gls{mi}. An existing input compression-based~\gls{ge} 
bound is used to link~\gls{mi} and~\gls{ge}. 
To the best of our knowledge, this represents the first empirical 
study of this bound. In our attempt to empirically falsify 
the theoretical bound, we find that it is often tight for 
best-performing models. Furthermore, it detects randomization 
of training labels in many cases, reflects test-time perturbation
robustness, and works well given only few training samples. 
These results are promising given that input compression is 
broadly applicable where~\gls{mi} can be estimated with confidence.

\end{abstract}

\glsresetall

\section{Introduction}

\gls{ge} is the central quantity for the performance assessment of~\glspl{dnn}, 
which we operationalize as the difference between the train-set accuracy 
and the test-set accuracy\footnote{\gls{ge} is also referred to
as~\emph{generalization gap}. Note that some use 
``generalization error'' as a synonym for ``test error''.}.
Bounding a~\gls{dnn}'s~\gls{ge} based on a training set is a 
longstanding goal~\citep{jiang2021methods} for various reasons: 
i) Labeled data is often 
scarce, making it at times impractical to set aside a representative test set. 
ii) The ability to predict generalization is expected to yield overarching 
design principles that may be used for~\gls{nas}, reducing a reliance on
trial-and-error. iii) Bounding the error rate is helpful for 
model comparison and essential for establishing performance guarantees for 
safety-critical applications. 
In contrast, the test accuracy is merely a single performance estimate based 
on an arbitrary and finite set of examples. 
Furthermore, the~\emph{adversarial examples} phenomenon 
has revealed the striking inability of~\glspl{dnn} to generalize in the presence of 
human-imperceptible perturbations~\citep{szegedy2014intriguinga, biggio2018wild},
highlighting the need for a more specific measure of~\emph{robust} generalization.

Various proxies for \gls{dnn} complexity which are assumed to be relevant 
to~\gls{ge}---such as network depth, width, $\ell_p$-norm bounds~\citep{neyshabur2015normbased}, 
or number of parameters---do not consistently predict generalization in
practice~\citep{zhang2021understanding}.
In search of an effective measure to capture the~\gls{ge} across a 
range of tasks, we investigate the \gls{mi} between the input and final layer 
representations, evaluated solely on the training set. 
In particular, we empirically study the 
\gls{icb} introduced by~\citep{tishby2017information, shwartz-ziv2019representation}, 
linking~\gls{mi} and several~\gls{ge} metrics. 
An emphasis on~\emph{input} is an important distinction from many 
previously proposed~\gls{ge} bounds (e.g.,~\cite{zhou2019nonvacuousa}), 
which tend to be \textit{model}-centric rather than \textit{data}-centric.

We use~\emph{infinite ensembles of infinite-width networks}~\citep{lee2019wide} for 
which~\gls{mi} is well defined. These models correspond to \textit{kernel regression} 
and are simpler to analyze than finite-width~\glspl{dnn}, yet they exhibit 
double-descent and overfitting phenomena observed in deep learning~\citep{belkin2019reconciling}.
For these reasons, \citet{belkin2018understand} suggested that understanding kernel 
learning should be the first step taken towards understanding generalization in deep learning. 
To this end, we evaluate the~\gls{icb} proposed by~\cite{tishby2017information, shwartz-ziv2019representation} 
with respect to three axes of performance:
\begin{enumerate}

\item First, we try to empirically~\emph{falsify} the bound by evaluating
the~\gls{ge} of a variety of models, composed by drawing random metaparameters
of the neural architecture and training procedure. 
We then compare the empirical~\gls{ge} to the theoretical~\gls{ge} bound 
given by~\gls{icb}. 
We show that~\gls{icb} contains the~\gls{ge} at the expected \(95\%\) confidence 
level for three of five datasets, or all five for the best-performing models. 
In addition, we suggest the training-label randomization
test~\citep{zhang2017understanding} as a means to determine when ICB may 
perform well a priori without relying on a test set.

\item Next, we analyze whether the~\gls{icb} is sufficiently small for useful model 
comparisons. If a theoretical~\gls{ge} bound exceeds \(100\%\) in practice, it is 
said to be~\emph{vacuous}. As we study binary classification tasks we additionally 
require that the bound be less than \(50\%\) for models with non-trivial~\gls{ge}.
We show that~\gls{icb} is often sufficiently close to the empirical~\gls{ge}, 
and thus presents a~\emph{non-vacuous} bound, obtained from less than 2000 
training samples. 

\item Last, we assess the~\emph{correlation} between~\gls{icb} and~\gls{ge}. 
Ranking~\gls{ge} is less consistent when several metaparameters vary, 
with~\gls{icb} sometimes outperforming, and at times under-performing a simpler 
baseline. Increasing the~\gls{ntk} diagonal regularization coefficient is 
most correlated with reducing~\gls{icb}.

\end{enumerate}

Beyond these three main desiderata for generalization bounds, 
we show advantages in reducing~\gls{icb} even when the~\gls{ge} is small. Reducing~\gls{icb} on~\emph{natural} 
training labels prevents models from fitting~\emph{random}
labels, and conversely, \gls{icb}~\emph{increases} when models are trained 
on~\emph{random} versus~\emph{natural} training labels~\citep{zhang2017understanding, zhang2021understanding}. 
Finally, we show that~\gls{icb} is predictive of test-time perturbation 
robustness~\citep{goodfellow2015explaining, gilmer2019adversarial}, without 
assuming access to a differentiable model. 

\biblio

\section{Background}
\label{sec:background}

We make use of an information-theoretically motivated generalization bound, 
the \gls{icb}, to establish an overlooked link between~\gls{mi} and~\gls{ge}. 
The bound seems to have first appeared in a lecture series (see,
e.g.,~\cite{tishby2017information}), later in a 
pre-print~\citep{shwartz-ziv2019representation}[Thm.~1] and more 
recently in a thesis~\citep{shwartz-ziv2022information}[Chapter 3].
Although an information-theoretic proof was provided, to the best 
of our knowledge the bound has not yet been studied empirically.

\subsection{Mutual information in infinite-width networks}
\label{sec:mi-ntk}

The~\gls{mi} between two random variables \(X\) and \(Z\) is 
defined as 
\begin{equation}\label{eq:MI}
    I(X; Z)\equiv \sum_{x,z} p(x, z) \log\frac{p(x, z)}{p(x)p(z)} 
    = \mathbb{E}_{p(x, z)} \left[ \log \frac{p(z | x)}{p(z)} \right],
\end{equation}
where we used Bayes' rule to obtain the expression on the right and 
introduced \(\mathbb{E}_{p(x, z)}[\cdot]\) to denote the average over \(p(x, z)\). 
In our case, \(X\) denotes the input and \(Z\) the input representation, i.e., the output of any of the network's layers.  
Since the marginal \(p(z)\) is unknown, we use an unnormalized multi-sample 
``noise contrastive estimation'' (InfoNCE) variational bound. 
The InfoNCE procedure was originally proposed for unsupervised representation
learning~\citep{oord2018representation}, which also serves as a lower bound
on~\gls{mi}~\citep{poole2019variational}.
In~\cite{oord2018representation}, the density ratio \( p(z|x) / p(z) \) was 
learned by a neural network. Instead, 
following~\cite{shwartz-ziv2020information}, we use infinite ensembles 
of infinitely-wide neural networks, which have a conditional Gaussian 
predictive distribution: 
\( p(z | x) \sim \mathcal{N}\left(\mu(x, \tau), \Sigma(x, \tau)\right) \)
with \(\mu, \Sigma\) given by the~\gls{ntk} and Neural Network Gaussian Process 
(NNGP) kernel~\citep{jacot2018neural}.
The predictive distribution also remains Gaussian following \(\tau\) 
steps of~\gls{gd} on the~\gls{mse} loss.
The conditional Gaussian structure given by~\gls{ntk} may be supplied in the 
InfoNCE procedure, yielding~\gls{mi} bounds free from variational parameters. 
Specifically, we use the ``leave one out'' upper bound~\citep{poole2019variational} 
on~\gls{mi} to conservatively bound~\gls{mi}:
\begin{equation} \label{eq:ub}
I(X; Z) \leq \E \left[ \frac{1}{N} \sum_{i=1}^{N} \log
\frac{p(z_i | x_i)}{\frac{1}{N-1} \sum_{j \neq i} p(z_i | x_j)} \right] = I_{\text{UB}} .\
\end{equation}
A lower bound on~\gls{mi}, \(I_\text{LB}\), of a similar form as~\eqref{eq:ub} 
is also available (\eqref{eq:lb}, Appendix~\ref{sec:sub-appendix-lower-bound-mi}),
and we verified that both bounds yield similar results for the training regime 
in which we apply them (Fig.~A\ref{fig:euro-teaser-mi}). 
Note that since we wish to evaluate a \emph{generalization} bound, it is 
important that these~\gls{mi} bounds are computed on the training set only.

\subsection{Input compression bound}
\label{sec:icb}

Here, we provide an intuitive explanation of the~\gls{icb}
building on existing results and using information theory fundamentals~\citep{cover1991elements}. 
A more formal derivation including a proof can be found in~\cite{shwartz-ziv2019representation}[Appendix A]. 
We begin with the conventional GE bound developed in the PAC framework, 
which plays a central role in the early mathematical descriptions of machine learning. 
The abbreviation PAC stands for {\it Probably Approximately Correct} and provides a 
probabilistic formulation for the learning success. It is assumed that a model 
receives a sequence of examples $x$, each labeled with the value $f(x)$ of a 
particular target function, and has to select a hypothesis that approximates $f$ 
well from a certain class of possible functions. 
By relating the hypothesis-class cardinality \(|\mathcal{H}|\), the confidence 
parameter $\delta$, and the number of training examples \(N_\text{trn}\), one obtains 
the following bound on the GE: 
\begin{equation}\label{eq:GG}
    \mathrm{GE} < \sqrt{ \frac{\log(|\mathcal{H}|)+\log(1/\delta)}{2 N_\text{trn}} }\,.
\end{equation}
The key term in this bound is the hypothesis-class cardinality, 
which is essentially the \textit{expressive power} of the chosen ansatz. 
For a finite \(\mathcal{H}\), it is simply the number of possible functions 
in this class; when \(\mathcal{H}\) is infinite, a discretization procedure 
is applied in order to obtain a finite set of functions.
For neural networks, \(|\mathcal{H}|\) is usually assumed to increase with 
the number of trainable parameters. The bound~(\ref{eq:GG}) states that
generalization is only possible when the expressivity is outweighed by 
the size of the training set, reflecting the well-known bias-variance trade-off 
of statistical learning theory.
Empirical evidence, however, demonstrates that this trade-off is qualitatively
different in deep learning, where generalization tends to improve as the neural
network size increases even when the size of the training set is held constant.

The key idea behind the ICB is to shift the focus from the hypothesis to the
\textit{input space}. 
For instance, consider binary classification where each of the \(|\mathcal{X}|\) 
inputs belongs to one of two classes. The approach that leads to bound~(\ref{eq:GG}) 
reasons that there are \(2^{|\mathcal{X}|}\) possible label assignments, only one of 
which is true, and hence a hypothesis space with \(2^{|\mathcal{X}|}\) Boolean
functions is required to guarantee that the correct labeling can be learned. 
The implicit assumptions made here are that all inputs are fully distinct and 
that all possible label assignments are equiprobable. 
These assumptions do not hold true in general, since classification fundamentally
\textit{relies} on similarity between inputs. However, the notion of similarity is 
data-specific and a priori unknown; thus, the uniformity assumption is required 
when deriving a general statement. 

The approach behind ICB circumvents these difficulties altogether by applying
information theory to the process of neural-network learning. 
First, note that solving a classification task involves finding a suitable 
partition of the input space by class membership.
Neural networks perform classification by creating a representation \(Z\) for 
each input \(X\) and progressively coarsening it towards the class label, 
which is commonly represented as an indicator vector. 
The coarsening procedure is an inherent property 
of the neural-network function, which is implicitly contained in \(Z\). 
By construction, the NN implements a partitioning of the input space, 
which is adjusted in the course of training to reflect the true class membership. 
In this sense, the cardinality of the hypothesis space reduces to
\(|\mathcal{H}|\approx 2^{|\mathcal{T}|}\),
where \(|\mathcal{T}|\) is the number of class-homogeneous clusters that
the~\gls{nn} distinguishes. 
To estimate \(|\mathcal{T}|\), the notion of {\it typicality} is employed: 
\textit{Typical} inputs have a Shannon entropy \(H(X)\) that is roughly equal 
to the average entropy of the source distribution and consequently a probability 
close to \(2^{-H(X)}\). 
Since the typical set has a probability of nearly \(1\), we can estimate 
the size of the input space to be approximately equal to the size of the 
typical set, namely $2^{H(X)}$. 
Similarly, the average size of each partition is given by $2^{H(X|Z)}$. 
An estimate for the number of clusters can then be obtained by 
$|\mathcal{T}| \approx 2^{H(X)}/2^{H(X|Z)} = 2^{I(X; Z)}$, 
yielding a hypothesis class cardinality $|\mathcal{H}| \approx 2^{2^{I(X; Z)}}$.
With this, the final expression for the ICB is:
\begin{equation}\label{eq:ICB}
\mathrm{GE_{ICB}} < \sqrt{ \frac{ 2^{I(X; Z)} + \log(1/\delta) }{ 2 N_\text{trn} }}\,.
\end{equation}
We only evaluate~\gls{icb} when we can obtain a confident estimate of \(I(X; Z)\).  
For this we require a tight sandwich bound on \(I(X; Z)\) with \(I_\text{UB} 
\approx I_\text{LB}\). 
We discard samples where \(I_\text{UB}(X; Z) > \log(N_\text{trn})\),  since 
\(I_\text{LB}(X; Z)\) cannot exceed \(\log(N_\text{trn})\). 
See Fig.~A\ref{fig:euro-teaser-mi} for typical \(I_\text{UB}, I_\text{LB}\) 
values during training and samples to discard. Note that the units 
for \(I(X; Z)\) in~\gls{icb} are~\emph{bits}.

\biblio

\section{Experiments}
\label{sec:experiments}

Our experiments are structured around three key questions: 1) Can we
falsify~\gls{icb}, e.g., by finding models with~\gls{ge} that exceeds the 
theoretical bound (\S\ref{sec:sub-res-bounding-ge}), 
or by falsely predicting generalization when randomizing the training labels
(\S\ref{sec:sub-exp-randomization})?
2) Is the ICB close enough to the empirical GE for useful model 
comparisons (\S\ref{sec:sub-res-vacuous-ge})? 
3) To what extent does~\gls{icb} correlate with~\gls{ge} evaluated 
on standard and robust test sets (\S\ref{sec:sub-res-ranking})?
Here, we describe the two main experimental procedures, Exp.~A 
(\S\ref{sec:sub-exp-through-training}) and Exp.~B (\S\ref{sec:sub-exp-steady}), 
in which we draw a population of models for comparison of~\glspl{ge} 
to the theoretical~\gls{icb}.

We focus on binary classification like much of the generalization 
literature, which also enables us to more efficiently evaluate~\gls{mi} 
bounds by processing kernel matrices that scale by 
\(N_\text{trn}^2\) rather than \((k \times N_\text{trn})^2\) for \(k\)-classes. 
Aside from this computational advantage, our methodology extends 
to the multi-class setting.

\subsection{Evaluating generalization throughout training (Experiment A)}
\label{sec:sub-exp-through-training}
 
We conduct experiments with five standard benchmark datasets:
\texttt{MNIST}~\citep{lecun1998mnist},~\texttt{FashionMNIST}~\citep{xiao2017fashionmnist},
\texttt{SVHN}~\citep{netzer2011reading}, \texttt{CIFAR-10}~\citep{krizhevsky2009learning}, 
and~\texttt{EuroSAT}~\citep{helber2018introducing, helber2019eurosat}.
These datasets are intended to be representative of low to moderate complexity tasks
and make it tractable to train thousands of models~\citep{jiang*2019fantastic}. 
Experiments with~\texttt{EuroSAT} further demonstrate how the method scales to 
64-by-64 pixel images.
For each of the image datasets, we devise nine binary classification tasks 
corresponding to labels ``$i$ versus $i+1$'' for \(i \in \{0, \dots, 8\}\). 
Note that this sequential class ordering is an arbitrary choice.

We initialize a variety of models by sampling uniformly at random from 
the following metaparameters: 
the number of fully-connected layers, \(L \sim \mathcal{U}(1, 5)\), 
the diagonal regularization coefficient \(\lambda \sim \mathcal{U}\{10^0, 10^{-1},
10^{-2}, 10^{-3}, 10^{-4}\}\), 
the activation function \(\phi(\cdot) \sim \mathcal{U}\{ \texttt{ReLU}(\cdot),
\texttt{Erf}(\cdot) \} \), and 
the number of training samples, \(N_\text{trn} \sim \mathcal{U}(250, 2000)\). 
Test sets have a constant size of \(N_\text{tst} = 2000\). 
We do not randomly sample a learning rate or mini-batch size, as the
infinite-width networks are trained by full-batch~\gls{gd}, 
for which the training dynamics do not depend on the learning rate once below a 
critical stable value~\citep{lee2019wide}. A nominal learning rate of 1.0 was 
used in all cases and found to be sufficient.\footnote{This was the default setting 
in~\texttt{neural\_tangents} software library~\citep{novak2020neural}.} 
We use 100 different random seeds to draw metaparameters 
for each of the nine tasks, yielding 900 models for each dataset.

Each of these 900 models was evaluated at five different time steps throughout 
training at $t = \{10^2, 10^3, 10^4, 10^5, 10^6\}$ yielding 4500 tuples 
(\gls{icb}, \gls{ge}) to analyze. 
The end points $t=10^2$ and $t=10^6$ were selected as most of the variation 
in~\gls{ge} was contained within this range. 
Training for less than $t=10^2$ steps typically resulted in a small~\gls{ge}, 
as both training and test accuracy were near random chance or increasing in lockstep.
In terms of steady-state behaviour,~\gls{ge} was often stable beyond $t=10^6$. 
Furthermore, $t=10^6$ was found to be a critical time beyond which $I_\text{UB}(X; Z)$ 
sometimes exceeded its upper confidence limit of $\log(N_\text{trn})$, 
particularly for small $\lambda$ values where memorization (lack of compression) is possible.

\subsection{Evaluating generalization at steady state (Experiment B)}
\label{sec:sub-exp-steady}

Binary classification tasks were devised from the same source datasets 
as in~\S~\ref{sec:sub-exp-through-training}. Instead of considering only nine tasks,
we enumerated all \({\binom{10}{2}}=45\) binary label combinations. 
For example for~\texttt{MNIST}, the classification task of distinguishing digit ``0'' 
versus ``1'', ``0'' versus ``2'', and so forth. 
Here, we used a fixed \(N_\text{trn}=1000\) for~\texttt{MNIST}, \texttt{FashionMNIST}, 
and~\texttt{EuroSAT}; and \(N_\text{trn}=2000\) for~\texttt{SVHN} and~\texttt{CIFAR-10}. 
We perform a uniform random search over: 
the number of fully-connected layers, $L \sim \mathcal{U}(1, 5)$, 
diagonal regularization coefficient, $\lambda \sim \mathcal{U}(0, 2)$, and 
activation function, $\phi(\cdot) \sim \mathcal{U}\{ \texttt{ReLU}(\cdot), \texttt{Erf}(\cdot) \}$.
We use 100 different random seeds to draw metaparameters for each of the 45 tasks, 
yielding 4500 trials for each source dataset. Each of the trials was evaluated 
at \(t = \infty\) yielding 4500 tuples (\gls{icb}, \gls{ge}).

\biblio

\section{Results}

\begin{figure}
\centering
\includegraphics[width=\textwidth]{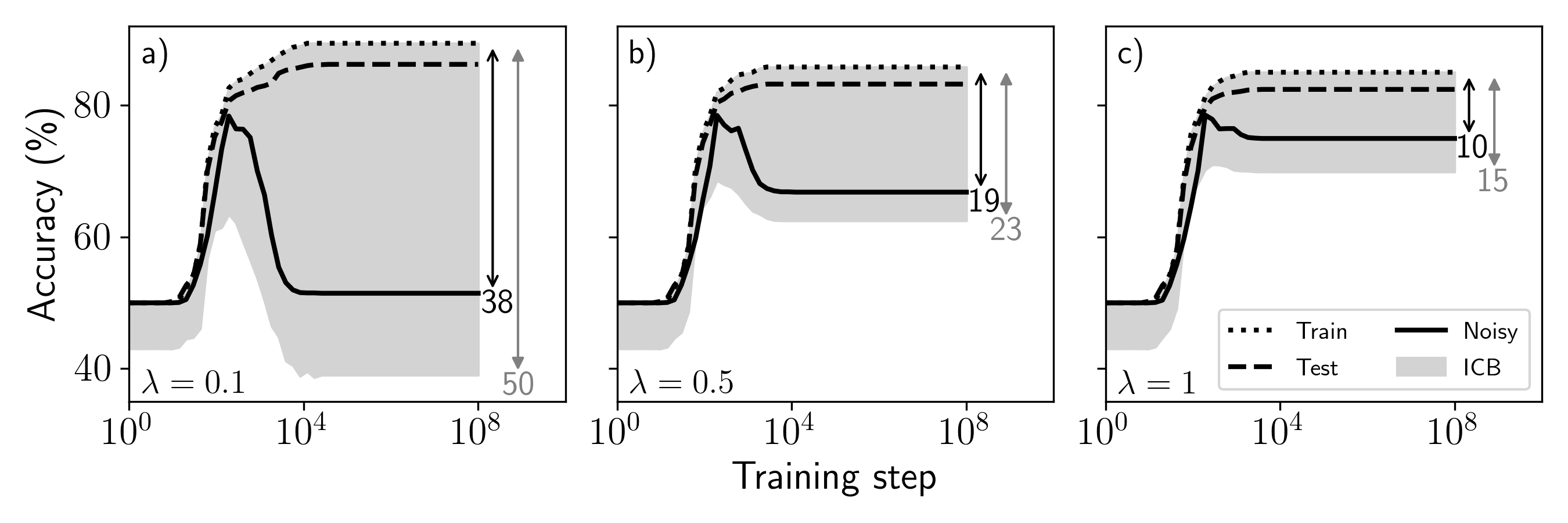}
\caption{The~\gls{icb} ({\color{gray} grey} filled area) evaluated on the~\texttt{EuroSAT}~\texttt{Pasture} versus~\texttt{Sea-Lake} binary classification task with 500 training 
samples and 2000 test samples at different regularization levels. At low regularization 
\(\lambda=0.1\) (a), the~\gls{icb} is vacuous with respect to standard generalization 
beyond $10^4$ training steps, but reflects the poor robust generalization for the 
``Noisy'' test set with~\gls{awgn}. 
Increasing the regularization to \(\lambda=0.5\) (b) and \(\lambda=1.0\) (c) 
reduces~\gls{icb} and the~\gls{awgn}~\gls{ge}. Arrows indicate the steady-state~\gls{awgn} \gls{ge} 
(black), and~\gls{icb} ({\color{gray} grey}) along with their respective values. 
See Fig.~A\ref{fig:euro-teaser-mi} for the corresponding upper and lower \(I(X; Z)\) 
bounds for this experiment.}
\label{fig:euro-teaser}
\end{figure}

\paragraph{Illustrative example}
Before presenting the main results, we examine~\gls{icb} for
a~\texttt{EuroSAT} classification task using only \(500\) training samples 
(Fig.~\ref{fig:euro-teaser}).
This is a challenging task, as tight~\gls{mi} and~\gls{ge} bounds are 
difficult to obtain for high-dimensional~\glspl{dnn}, particularly with few samples.
For example, in~\citep{dziugaite2017computing} \(55000\) samples were used to obtain 
a \(\approx 20\%\)~\gls{ge} bound for finite-width~\glspl{dnn} evaluated on~\texttt{MNIST}.

We evaluate~\gls{icb} throughout training from the first training step (\(t=10^0\)) 
until steady state when all accuracies stabilize (\(t=10^8\)). 
Shortly after model initialization ($t=10^0$ to $t=10^1$) the~\gls{icb} 
is $<7\%$ (indicated by the height of the shaded region in Fig.~\ref{fig:euro-teaser}) 
and the training and test accuracy are both at $50\%$ (\gls{ge}\(=0\)). 
Here,~\gls{icb} is non-vacuous, but also not necessarily interesting 
for this random-guessing phase. 
\gls{icb} increases as training is prolonged.\footnote{It may not be obvious 
that ICB increases monotonically with training steps as the 
training accuracy also increases.}
At low regularization (Fig.~\ref{fig:euro-teaser} a), the~\gls{icb} 
ultimately becomes vacuous (ICB $\approx 50\%$) around $10^4$ steps.
However, although~\gls{icb} is vacuous with respect to~\emph{standard} generalization in a), 
it reflects well the poor~\emph{robust} generalization when tested with~\gls{awgn}~\citep{gilmer2019adversarial}. 
Increasing the regularization coefficient \(\lambda\) reduces~\gls{icb} 
from $50\%$ (a) to $23\%$ (b) and $15\%$ (c), and the robust \gls{ge} 
from $38\%$ (a) to $19\%$ (b) and $10\%$ (c). 

Both standard and robust~\gls{ge} are bounded at all times by~\gls{icb}. 
The latter is, however, a coincidence, as the robust~\gls{ge} is subject to the 
arbitrary~\gls{awgn} noise variance ($\sigma^2 = \nicefrac{1}{16}$). 
The additive noise variance could be increased to increase the robust~\gls{ge} 
beyond the range bounded by~\gls{icb}.
More important than bounding the robust~\gls{ge}~\emph{absolute percentage} is 
that~\gls{icb} captures the trend of robust generalization. Evaluating 
robustness effectively is error-prone and often assumes access to test data 
and a differentiable model~\citep{athalye2018obfuscated, carlini2019evaluating}.
In contrast, we make no such assumptions here. The lack of robustness in 
Fig.~\ref{fig:euro-teaser} a) would have likely gone unnoticed by the practitioner. 
Either early stopping or increasing \(\lambda\) reduce the~\gls{icb} 
and robust~\gls{ge} as a potential solution---or a better trade-off 
between accuracy versus robustness~\citep{tsipras2019robustness}. 
A caveat to this example is that only two metaparameters varied: the number of 
training steps \(t\) and regularization \(\lambda \). 
Next, we assess the ability of~\gls{icb} to bound and rank~\gls{ge} for a broader 
range of metaparameters and datasets.

\biblio


\subsection{Bounding generalization error}
\label{sec:sub-res-bounding-ge}

\begin{figure}
\centering
\includegraphics[width=.95\textwidth]{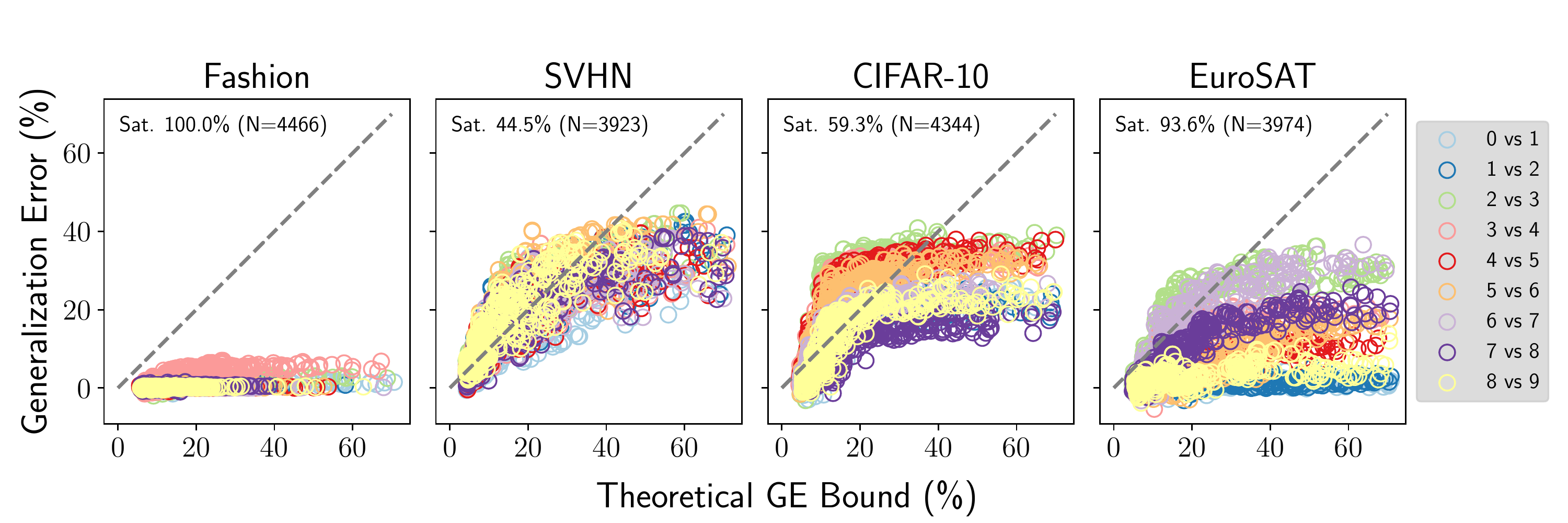}
\caption{The~\gls{icb} (``Theoretical GE Bound'') is plotted 
versus~\gls{ge} for~\texttt{FashionMNIST}, \texttt{SVHN}, 
\texttt{CIFAR}, and \texttt{EuroSAT} datasets for 
Exp.~A (\S\ref{sec:sub-exp-through-training}). 
We refer to the percentage of models with~\gls{ge} \(<\)~\gls{icb} 
as the~\gls{icb} satisfaction rate, 
which is annotated in the top left corner of each plot with format 
``Sat.~\% (N)'', where N denotes the number of valid experiments. 
Each binary classification task is assigned a unique colour to highlight inter-task 
differences in~\gls{icb} satisfaction rate.} 
\label{fig:bound_ge_acc_throughout_training}
\end{figure}

We refer to the percentage of tuples (\gls{icb}, \gls{ge}) for which \gls{ge} < \gls{icb}
as the \emph{``\gls{icb} satisfaction rate''}, or \emph{``Sat.''} in plots. 
We expect \(\approx 95\%\) of samples to satisfy this property as the bound 
is evaluated with \(\delta=0.05\) or \(95\%\) confidence. 
The overall~\gls{icb} satisfaction rate with the respective number of valid 
experiments \(N\) is listed in Table~\ref{tab:icb-sat-overview}. 
Exp.~B yielded greater \(N\) than Exp.~A primarily because it uses a different 
range for the regularization coefficient \(\lambda\), resulting in larger \(\lambda\)
values. Since larger \(\lambda\) is associated with more compression, Exp.~B had 
fewer samples being discarded than in Exp.~A 
due to \(I_\text{UB}\) exceeding \(\log(N_\text{trn})\). Otherwise, 
\gls{icb} satisfaction rates are similar, with \texttt{SVHN} 
performing slightly worse and \texttt{CIFAR-10} slightly better for 
Exp.~B versus Exp.~A. These results also suggest that exploring nine binary 
classification tasks (Exp.~A) serves as a useful approximation for the full 
set of all \(45\) possible tasks (Exp.~B).
Next, we analyze how model performance influences the~\gls{icb} satisfaction rate.

\begin{table}[]
\centering
\caption{Overall \gls{icb} satisfaction rate (Sat.~\%) with number of valid 
experiments \(N\) in brackets. Results for Exp.~A are also plotted in
Figure~\ref{fig:bound_ge_acc_throughout_training}; a more detailed breakdown 
of these results can be found in Table A\ref{tab:mnist-detailed}-\ref{tab:eurosat-detailed}.}
\begin{tabular}{l|lllll}
\toprule
& \texttt{MNIST} & \texttt{Fashion} & \texttt{SVHN} & \texttt{CIFAR} & \texttt{EuroSAT} \\ \midrule
\textbf{Exp.~A} 
& \(100\%~(2237)\) 
& \(100\%~(4466)\) 
& \(44.5\%~(3923)\) 
& \(59.3\%~(4344)\)
& \(93.6\%~(3974)\) \\
\textbf{Exp.~B} 
& \(100\%~(2250)\)
& \(100\%~(4500)\)
& \(27.0\%~(4500)\)
& \(68.0\%~(4500)\)
& \(95.0\%~(2221)\) \\
\bottomrule
\end{tabular}
\label{tab:icb-sat-overview}
\end{table}

When we restrict our scope to the best-performing models based on their 
test accuracy, the \gls{icb} satisfaction rate improves considerably. 
For example, models with test accuracy \(\geq 80\%\) attain 
\gls{icb} satisfaction rates of \(94\%~(N=682)\) for \texttt{SVHN} in 
Exp.~B, and \(99\%~(N=2812)\) for \texttt{EuroSAT} in Exp.~A (Fig.~A\ref{fig:eurosat-v4-test-acc-80}). 
For \texttt{CIFAR-10} in Exp.~B, we obtain \(96\%~(N=591)\) by restricting to test accuracy \(\geq 87\%\).
The specific test accuracy thresholds were chosen to balance a trade-off 
between satisfying the~\gls{icb} at \(\approx 95\%\) and maximizing \(N\). 
Although best-performing models are more likely to be deployed in practice, 
theoretical \gls{ge} bounds generally prohibit access to a test set. 
Therefore, we next select models based on their \textit{training} accuracy.

\begin{wrapfigure}{r}{0.36\textwidth}
\centering
\includegraphics[width=0.35\textwidth]{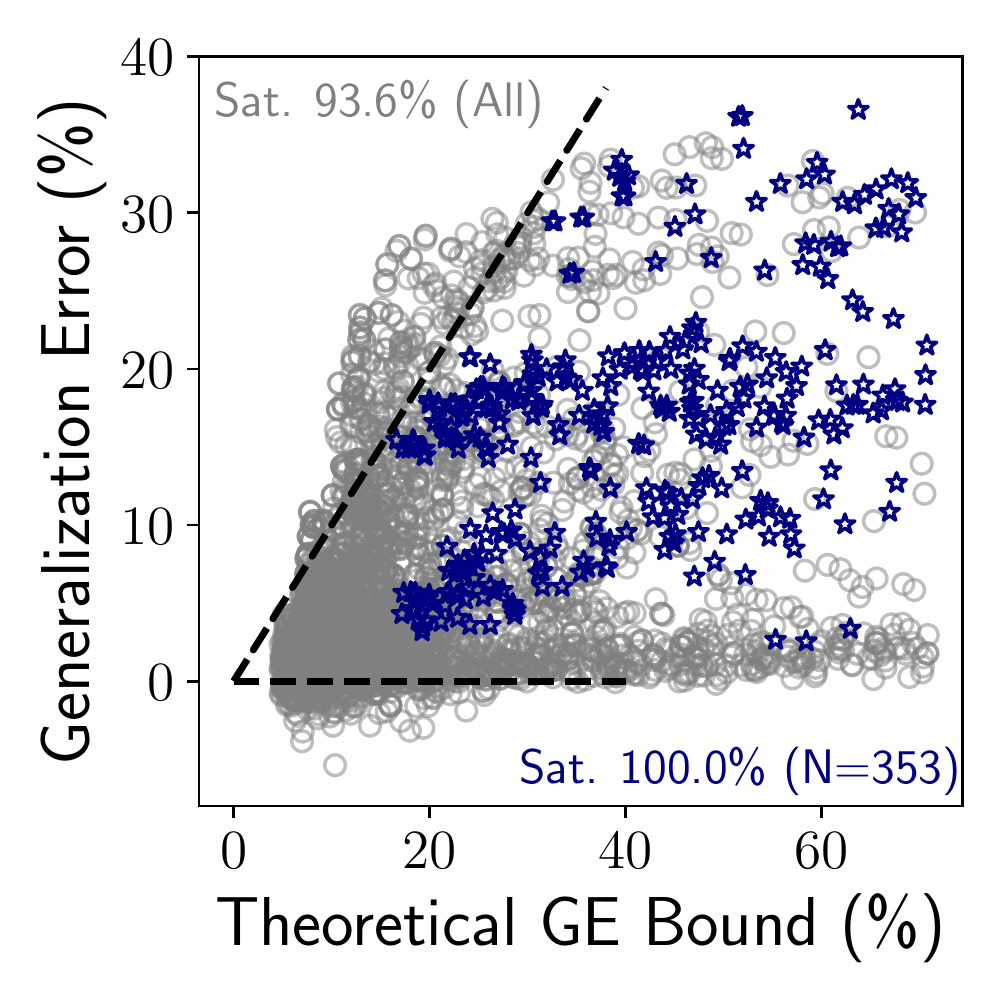}
\caption{Theoretical~\gls{ge} bound (ICB) versus~\gls{ge} for~\texttt{EuroSAT} 
(Exp.~A). Overfitted models indicated by stars.}
\label{fig:euro-overfitted}
\end{wrapfigure}

We refer to models that achieve \(100\%\) accuracy on the training set as 
``overfitted'', consistent with prior use of this term by~\citet{belkin2018understand}.
Interestingly, restricting our analysis to overfitted models either improves or 
does not change~\gls{icb} satisfaction rate for Exp.~A. 
For \texttt{MNIST}, \texttt{FashionMNIST}, \texttt{SVHN}, and
\texttt{EuroSAT}, overfitted models attain an \gls{icb} satisfaction rate of 
\(100\%\) with \(N=1970, 1820, 20, 353\) respectively, while for~\texttt{CIFAR-10}, 
the satisfaction rate remained below \(95\%\), albeit it improved 
from \(59.3\%~(N=4344)\) to \(72.6\%~(N=876)\). 
Similar results were observed for Exp.~B. 
Detailed plots for these results can be found in 
Figures~A\ref{fig:mnist-scatter}, A\ref{fig:fashion-scatter}, A\ref{fig:svhn-scatter},
A\ref{fig:cifar-scatter}, A\ref{fig:eurosat-scatter} in the Appendix. 
The mostly excellent~\gls{icb} satisfaction rates of the overfitted models 
are not due to trivially constant~\gls{ge} or~\gls{icb} values; 
as can be seen in Figure~\ref{fig:euro-overfitted}, 
these models still have considerable variance w.r.t.~both metrics 
despite their identical training accuracies.

Inter-task differences were observed in terms of the ability 
of~\gls{icb} to bound~\gls{ge}. For example for Exp.~A, six of nine~\texttt{EuroSAT} 
binary classification tasks~\emph{always} satisfied~\gls{icb} (\(N=2534)\), 
whereas two tasks in particular reduced the overall average. 
The satisfaction rate was only \(68\%~(N=468)\) for the ``2 vs.~3'' task 
and \(72\%~(N=475)\) for the ``6 vs.~7'' task (see Fig.~A\ref{fig:bound_ge_acc_throughout_training}
and Table A\ref{tab:eurosat-detailed} for detailed results).
These tasks were unusual in that there was a strong inverse relationship
between training error and~\gls{ge}, such that reducing the 
training error resulted in a steady increase in test error, with 
\(\tau \approx -0.9\) for both tasks, compared to \(\tau=-0.58\) for the 
``0 vs.~1'' task. The negative correlation between training and test performance
for the ``2 vs.~3'' task also resulted in a lower mean test accuracy 
(\(70.2 \pm 2.2\%~(N=467)\)) compared to other tasks, e.g., 
``0 vs.~1'' (\(93.5 \pm 5.3\%~(N=415)\)), consistent with our 
previous observation that best-performing models generally satisfy~\gls{icb}. 
We further investigated inter-task differences for~\texttt{EuroSAT} Exp.~B, 
for which all \({\binom{10}{2}}=45\) binary classification tasks were 
evaluated for 50 seeds each. For the two poorly performing tasks 
``2 vs.~3'' and ``6 vs.~7'', the~\gls{icb} satisfaction rate was \(78\%~(N=50)\) 
and \(86\%~(N=50)\), respectively. 
For 34 of 45 tasks \((N=1913)\), \gls{icb} was satisfied for all seeds.

\biblio


\subsection{The randomization test}
\label{sec:sub-exp-randomization}

\citet{zhang2017understanding, zhang2021understanding} proposed 
the ``randomization test'' 
after observing that~\glspl{dnn} easily fit random labels. They argue that 
useful generalization bounds ought to be able to distinguish models trained 
on~\emph{natural} versus~\emph{randomized} training labels, since generalization 
is by construction made impossible in the latter case. However, we cannot 
necessarily expect a theoretical~\gls{ge} bound to exactly hold for models trained 
on random labels, since the training and test sets are no longer drawn from the 
same distribution. We therefore pose the following questions: 
Q1: \textit{To what extent does the~\gls{icb} correlate with 
the ability to fit random labels?} Q2: \textit{Can~\gls{icb} distinguish 
training sets with natural versus random labels?} 
To address Q1, we aim to find metaparameters that reduce~\gls{icb} 
and prevent models from fitting~\emph{random} training labels, 
while still permitting them to fit the~\emph{natural} training labels.
This, however, introduces a potential for confounding if the metaparameter
choice alone prevents the model from fitting random labels rather than~\gls{icb}\@. 
For Q2, we hold all metaparameters constant and observe whether~\gls{icb} 
changes for randomized training labels.

\begin{figure}
\centering
\includegraphics[width=0.87\textwidth]{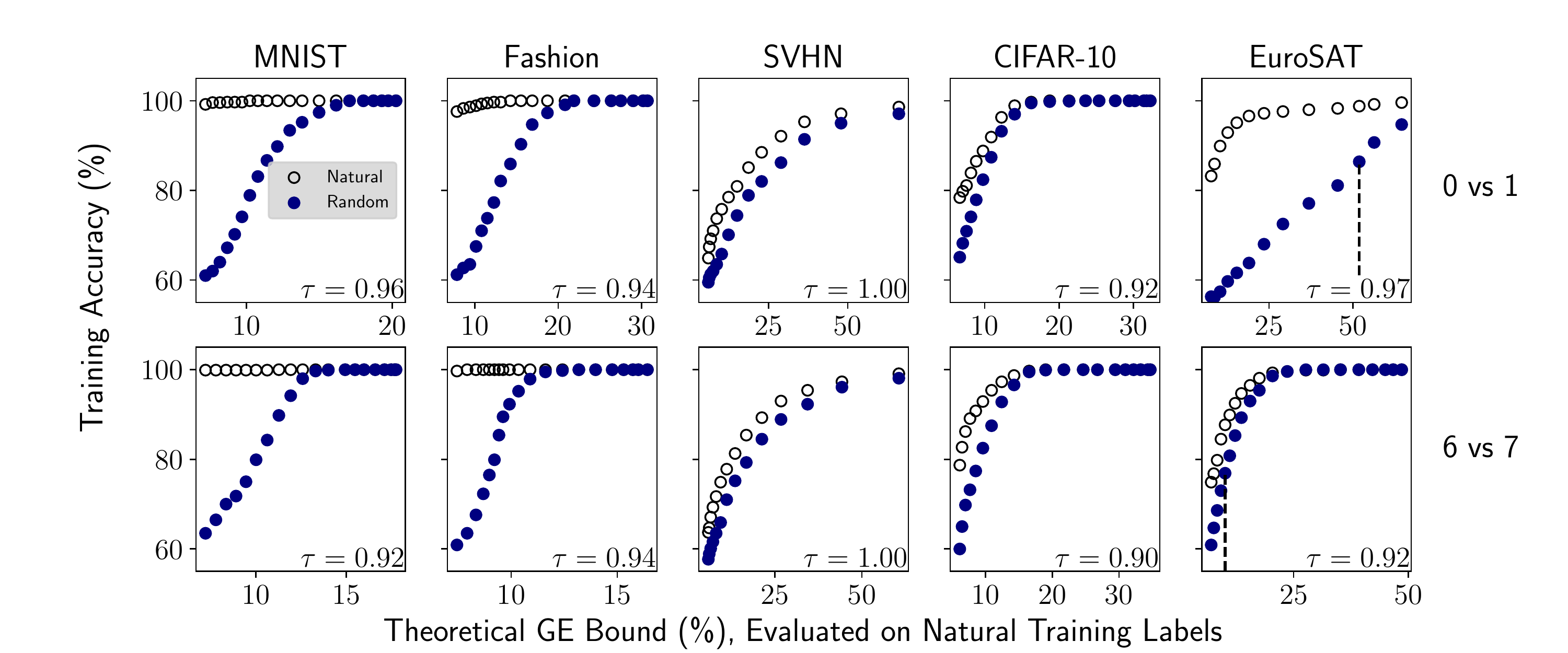}
\caption{\textbf{\gls{icb} often distinguishes between 
natural and randomized training sets.} The accuracy w.r.t.~the natural training 
labels (``Natural'') and randomized training labels (``Random'') is plotted 
versus the theoretical GE bound (\gls{icb}), which is evaluated on the natural 
training labels. Each data point in the plot corresponds to a unique regularization
value \(\lambda\), which in turn influences the~\gls{icb} value: Small \(\lambda\) leads to a large~\gls{icb} value, while increasing 
\(\lambda\) reduces the~\gls{icb} and the ability to fit random labels, with the aim 
of preserving accuracy for natural training labels. 
The top row corresponds to the ``0 vs.~1'' task and bottom row the ``6 vs.~7'' task. 
Considerable separation between natural and random labels 
for~\texttt{MNIST}, \texttt{FashionMNIST}, 
and~\texttt{EuroSAT} is observed. Differences are harder to distinguish for~\texttt{SVHN} 
and~\texttt{CIFAR-10}, but still apparent. 
\gls{icb} is highly correlated with ability to fit random labels in all cases.
The broken vertical line for~\texttt{EuroSAT} indicates the \gls{icb} value for 
which there is at least a 10\% difference in natural versus random training accuracy.}
\label{fig:rand-test}
\end{figure}

For simplicity, we consider a two-layer fully-connected ReLU network. 
We train the model to \(t=\infty\) on the natural training set 
(\(N_\text{trn}=1000\)) with 20 different regularization values 
\( \lambda \) in the range \( 10^{-4} \ \text{to} \ 10^{1} \). 
We measure Kendall's \(\tau\) ranking between the~\gls{icb} evaluated on these 
models and their training accuracies when re-trained with random labels. 
We find that the~\gls{icb} value obtained after training on~\emph{natural} labels 
is strongly correlated with the ability to fit~\emph{random} labels for all five 
datasets. Furthermore, competitive accuracy for~\emph{natural} 
training labels is preserved for three of five datasets in doing so
(see Fig.~\ref{fig:rand-test}).

\begin{wraptable}{r}{9.5cm}
\caption{\textbf{\gls{icb} is larger when training to identical accuracy 
on random versus natural labels.}
Detailed results of the randomization test for~\texttt{EuroSAT} are shown. 
The lower and upper~\gls{mi} bounds, \(I_\text{LB}\) and~\(I_\text{UB}\), are 
included for comparison against~\(\log(N_\text{trn}) \approx 6.91 \ \text{nats}\). 
The columns \( \text{ICB}_\text{LB} \) and \(\text{ICB}_\text{UB}\) refer 
to whether \(I_\text{LB}\) or \(I_\text{UB}\) is taken as \(I(X; Z)\) estimate, 
respectively. Columns ``Train'' and ``Test'' show the respective accuracy in \(\%\). 
Both ICB values are larger for the random case when comparing rows with ``Train''\(=100.0\).}
\centering
\begin{tabular}{rrrrrrrr}
\toprule
\multicolumn{8}{c}{Natural Training Labels} \\ \midrule
\( \lambda \) & \(I_\text{LB}\) & \(I_\text{UB}\) & \(\text{ICB}_\text{LB}\) & \(\text{ICB}_\text{UB}\) & Train & Test & GE \\ \midrule
\(10^{-1}\) & 4.87 & 5.37 & 26.0 & 33.1 & 97.9 & 98.7 & -0.8 \\
\(10^{-2}\) & 5.40 & 6.58 & 33.7 & 60.2 & 99.5 & 98.6 & 0.9 \\
\(10^{-3}\) & 5.78 & 7.40 & 40.5 & 90.4 & 100.0 & 97.5 & 2.5 \\ \cmidrule(l){1-8} 
\multicolumn{8}{c}{Random Training Labels} \\ \cmidrule(l){1-8} 
\(10^{-1}\) & 3.68 & 3.75 & 15.0 & 15.5 & 71.3 & 50.0 & 21.3 \\
\(10^{-2}\) & 5.28 & 5.67 & 31.7 & 38.5 & 89.7 & 50.0 & 39.7 \\
\(10^{-3}\) & 6.37 & 8.96 & 54.1 & 197.6 & 100.0 & 50.0 & 50.0 \\ \bottomrule
\end{tabular}
\label{tab:euro-rand-test}
\end{wraptable}

Surprisingly, \(\text{ICB}_\text{UB}\) approximates the~\gls{ge} well even 
when the model is trained on~\emph{random labels} (Table~\ref{tab:euro-rand-test}).
For \(\lambda = 0.1\), \(\text{ICB}_\text{UB} = 15.5\%\) compared to a~\gls{ge} of \(21.3 \% \).
Next, for \(\lambda = 0.01\), \(\text{ICB}_\text{UB} = 38.5\%\) and~\gls{ge} is \(39.7 \% \). 
Last, for \(\lambda = 0.001\), \(I_\text{UB} = 8.96\), which is greater than 
\(\log(N_\text{train}) = 6.91 \) nats, therefore the corresponding~\(\text{ICB}_\text{UB}\) 
of \(197.6\%\) should be discarded. In this case, substituting the ``optimistic'' lower
estimate \(\text{ICB}_\text{LB} = 54.1\% \approx \)~\gls{ge}~\(=50\%\).

Intuitively, we expect \(I(X; Z)\) to be smaller after training on natural labels,
since training on random labels requires memorization of random data, i.e., the 
opposite of compression. However, note that to isolate the effect of the
training label type on~\gls{icb}, the training accuracy must also be controlled, 
as higher accuracy generally requires greater complexity and thus larger \(I(X; Z)\). 
This intuition is consistent with our results, as both \(I_\text{LB}\) 
and \(I_\text{UB}\) increase monotonically with the training accuracy 
for both training label types (see Table~\ref{tab:euro-rand-test}).
    
Training with \(\lambda=0.001\) allows models to perfectly fit both natural and 
randomized training sets (Table~\ref{tab:euro-rand-test} column ``Train'' = \(100\%\)), 
which presents a suitable setting for evaluating whether~\gls{icb} is sensitive to 
whether training labels are natural or random.
Indeed, \( I_\text{LB} \) is greater for random labels (6.37 vs.~5.78 nats), 
resulting in an increase of the~\emph{optimistic} theoretical \gls{ge} bound, 
\(\text{ICB}_\text{LB}\), from \( 40.5\% \ \text{to} \ 54.1\% \). 
The more~\emph{pessimistic} \(\text{ICB}_\text{UB}\) increases 
even more dramatically from \( 90.4\% \ \text{to} \ 197.6\% \), which is 
beyond the valid range of~\gls{ge} (\(0-100\%\)).\looseness=-1

\paragraph{The randomization test identifies tasks with 
low~\gls{icb} satisfaction rate}

\begin{table}[]
\centering
\caption{\textbf{Ability of~\gls{icb} to separate natural versus random training labels 
is a good predictor of~\gls{icb} satisfaction rate by task.}
The row ``\(\text{ICB}_\text{rand} @ X\%\)'' indicates the minimum~\gls{icb} value for which a \(X\%\) accuracy 
difference between natural and random labels is observed. 
The ``Sat.~(\%)'' column showing the~\gls{icb} satisfaction rate is taken from \S\ref{sec:sub-res-bounding-ge}, Exp.~A. 
The column \(\tau\) indicates the rank correlation 
between \(\text{ICB}_\text{rand} @ X\%\) and Sat.~(\%) over the nine tasks. Columns 
sorted by ascending order of \(\text{ICB}_\text{rand} @ X\%\).}
\begin{tabular}{cccccccccccc}
\toprule
& Task                               & 2/3               & 6/7 &  3/4  & 7/8  & 4/5  & 5/6  & 0/1  & 1/2  & 8/9 & \\ 
\multirow{1}{*}{EuroSAT} & Sat.~\% & 68 & 72   & 97   & 100  & 100  & 100  & 100  & 100  & 100 & \multirow{2}{*}{\(\tau=0.76\)} \\
& \(\text{ICB}_\text{rand} @ 10\%\) & 7.5 & 10.0 & 11.2 & 13.4 & 19.3 & 22.5 & 51.9 & 59.5 & 66.9 \\ \bottomrule
& Task                                & 2/3                & 5/6 & 4/5 & 0/1 & 3/4  & 6/7  & 1/2  & 8/9 & 7/8 & \\ 
\multirow{1}{*}{CIFAR-10} & Sat.~\% & 29  & 41  & 39  & 74  & 35   & 68   & 79   & 74  & 97 & \multirow{2}{*}{\(\tau=0.65\)} \\
& \(\text{ICB}_\text{rand} @ 5\%\)    & 6.8 & 8.1 & 9.3 & 9.8 & 10.4 & 10.9 & 11.2 & 11.3 & 13.2 \\ \bottomrule
\end{tabular}
\label{tab:euro-rand-test-icb-sat}
\end{table}

Recall from \S\ref{sec:sub-res-bounding-ge} that three binary classification tasks were responsible 
for reducing the~\gls{icb} satisfaction rate below \(100\%\) for~\texttt{EuroSAT}: ``2 vs.~3'' (Sat.~68\%), ``6 vs.~7'' (Sat.~72\%), and ``3 vs.~4'' (Sat.~97\%) for Exp.~A.
We observed that these were the~\emph{same} tasks for which~\gls{icb} performed 
poorly on the randomization test. 
Specifically, we measured the minimum~\gls{icb} value for which a \(10\%\) or greater percentage 
difference was detected between the natural and random training-sets 
(see, e.g., the vertical broken line in Fig.~\ref{fig:rand-test}). 
The ``2 vs.~3'' task required the smallest~\gls{icb} (\(7.5\%\)) before the difference in 
label type became apparent. The ``6 vs.~7'' task had the next highest~\gls{icb} of \(10.0\%\), 
followed by ``3 vs.~4'' with \(11.2\%\). The other six tasks---that have \(100\%\) satisfaction 
rate---have strictly greater~\gls{icb} (Table~\ref{tab:euro-rand-test-icb-sat}). 
Similar results are observed for~\texttt{CIFAR-10} using a smaller \(5\%\) threshold as 
accuracies for natural and random labels were closer than for~\texttt{EuroSAT}. 
The tasks with minimum (``2 vs.~3'') and maximum (``7 vs.~8'') satisfaction rate are the same 
tasks with the minimum and maximum \(\text{ICB}_\text{rand} @ 5\%\).
Therefore, the training-set based randomization test---which only required 
training a single model here---may be used to help identify when~\gls{icb} 
performs well as a GE bound for a variety of models (e.g., the thousands of 
models that were considered in Exp.~A). 
Our adaptation of the well-known randomization test complements the list of 
factors already identified in \S\ref{sec:sub-res-bounding-ge} as affecting~\gls{icb} 
satisfaction rate.

\biblio


\subsection{Vacuous or non-vacuous?}
\label{sec:sub-res-vacuous-ge}

\begin{table}[]
\centering
\caption{\textbf{\gls{icb} is non-vacuous for best-performing models on five datasets.} 
The~\gls{ge} (\%) of a best-performing model is compared to~\gls{icb} for each dataset.
The column \(N_\text{trn}\) indicates the number of training samples,
which was a metaparameter for the first experiment (\S\ref{sec:sub-exp-through-training}, \textbf{Exp.~A}),
and a constant for the second experiment (\S\ref{sec:sub-exp-steady}, \textbf{Exp.~B}).}
\begin{tabular}{lrrrrr|rrrrr}
\toprule
 & \multicolumn{5}{c}{\textbf{Exp.~A}, \(t=\{10^2, \dots, 10^6\}\)} & \multicolumn{5}{c}{\textbf{Exp.~B}, \(t=\infty\)} \\ \midrule
Dataset & Train & Test & GE & ICB & \(N_\text{trn}\) & Train & Test & GE & ICB & \(N_\text{trn}\) \\ \midrule
\texttt{MNIST} & 100.0 & 100.0 & 0.0 & 11.2 & 931 & 99.9 & 99.9 & 0.0 & 12.1 & 1000 \\
\texttt{Fashion} & 99.9 & 100.0 & -0.1 & 7.2 & 1112 & 99.9 & 100.0 & -0.1 & 8.1 & 1000 \\
\texttt{SVHN} & 98.8 & 74.2 & 24.6 & 28.0 & 1564 & 100.0 & 90.8 & 9.3 & 21.1 & 2000 \\
\texttt{CIFAR} & 94.8 & 89.2 & 5.6 & 7.6 & 1966 & 99.2 & 93.8 & 5.4 & 11.3 & 2000 \\
\texttt{EuroSAT} & 97.8 & 98.7 & -0.9 & 25.6 & 1979 & 100.0 & 100.0 & 0.0 & 22.6 & 1000 \\ \bottomrule
\end{tabular}
\label{tab:nonvacuous}
\end{table}

Models with high test accuracy are the more likely to be deployed in practice. 
To evaluate whether~\gls{icb} is non-vacuous and close enough to~\gls{ge} to 
aid model comparison, for each dataset we selected the model with the 
smallest~\gls{icb} value among the top-three most accurate models. 
The~\gls{icb} values are considerably less than $50\%$ in all 
cases, satisfying the basic definition of non-vacuous for a binary 
classification task (Table~\ref{tab:nonvacuous}).
For Exp.~A, the smallest difference between~\gls{icb} and~\gls{ge} occurred for 
the~\texttt{CIFAR} dataset, with a~\gls{ge} of \(5.6\%\) compared to an~\gls{icb} 
value of \(7.6\%\) for \(1966\) training samples (Table~\ref{tab:nonvacuous}, Exp.~A). 
The greatest~\gls{icb} value occurred for~\texttt{SVHN} (\(28.0\%\)), 
however the~\gls{ge} was also large in this case (\(24.6\%\)). 

For Exp.~B, both~\gls{icb} and~\gls{ge} decrease for~\texttt{SVHN} 
(Table~\ref{tab:nonvacuous}, Exp.~B)) relative to Exp.~A. 
For~\texttt{CIFAR}, a similar~\gls{ge} 
of \(\approx 5\%\) is attained as in Exp.~A, but with a greater~\gls{icb} by \(3.7\%\).  
This may be due to the training accuracy increasing by \(4.4\%\) from 
\(94.8\%\)~(Exp.~A) to \(99.2\%\)~(Exp.~B). In summary, not only is 
ICB non-vacuous, it is close enough to~\gls{ge} to perform model comparisons.



\subsection{Relationship between theoretical bound and generalization error}
\label{sec:sub-res-ranking}

Here, we evaluate the ability of~\gls{icb} to rank~\glspl{ge} in terms 
of Kendall's rank correlation coefficient, \(\tau\).
Our analysis of correlation between a complexity metric and 
empirical~\gls{ge} is inspired by a previous study and
NeurIPS competition~\citep{jiang*2019fantastic, jiang2021methods}.  
Figure~A\ref{fig:mi-vs-icb} helps motivate the use of the~\gls{icb} 
for ranking~\gls{ge} rather than using its constituent complexity 
metric \(I(X; Z\)), based on a subset of Exp.~B metaparameters.

An issue with correlation analysis is that the training-set classification 
error or proxy loss can serve as a good predictor of~\gls{ge}, 
therefore~\citet{jiang*2019fantastic} train models to a fixed training loss to 
control for confounding effects. 
However, fixing the training loss limits the extent of metaparameter exploration.
For example, a complexity metric or GE bound may rank~\glspl{ge} of overfitted 
models well, but perform poorly for early-stopping.
To maintain a broad scope, we follow both Exp.~A \& B procedures and 
treat the train-set accuracy as a baseline for comparison against~\gls{icb}, 
then evaluate overfitted models separately. 

\begin{wraptable}{r}{7cm}
\centering
\caption{Kendall's $\tau$ ranking for three~\gls{ge} types: 
Clean, \gls{awgn} and \gls{fgsm} for models that obtain zero training error.
The number of models is indicated by the $N$ column. 
NB: The ``--'' entries for \texttt{SVHN} had \(p > 0.05\) 
when computing \(\tau\) and were therefore discarded.}
\label{tab:100-train}
\begin{tabular}{lrccc}
\toprule
Dataset & $N$ & Clean & AWGN & FGSM \\ \midrule
\texttt{MNIST} & 2329 & 0.27 & 0.30 & 0.29 \\
\texttt{Fashion} & 1820 & 0.39 & 0.42 & 0.41 \\
\texttt{SVHN} & 20 & 0.32 & -- & -- \\ 
\texttt{CIFAR} & 876 & 0.19 & 0.20 & 0.12 \\
\texttt{EuroSAT} & 353 & 0.33 & 0.38 & 0.29 \\ \bottomrule
\end{tabular}
\end{wraptable}

Two perturbed test sets help measure correlations between~\gls{icb} 
and~\emph{robust}~\gls{ge}; as perturbations we use~\gls{awgn}~\citep{gilmer2019adversarial}
and~\gls{fgsm}~\citep{goodfellow2015explaining}. These perturbations are appropriate 
for evaluating the robustness of infinite-width networks trained by~\gls{gd}, which behave as 
linear functions of their parameters~\citep{lee2019wide}.
It can be shown that a classifier's error rate for a test set corrupted 
by~\gls{awgn} determines the distance to the decision boundary for linear 
models~\citep{fawzi2016robustnessa} and serves as a useful guide
for~\glspl{dnn}~\citep{gilmer2019adversarial}. For~\gls{awgn}, we use a 
Gaussian variance \(\sigma^2=\nicefrac{1}{16}\) for~\texttt{EuroSAT} and 
\(\sigma^2=\nicefrac{1}{4}\) for the other datasets. 
For~\gls{fgsm}, we use a \(\ell_\infty\)-norm perturbation 
of size \(4/255\) for inputs \(x \in [-1, +1]\).

In terms of ranking (Clean, \gls{awgn}, \gls{fgsm})~\glspl{ge} by aggregating
all nine tasks for Exp.~A, \gls{icb} performs better than the training accuracy 
baseline for \texttt{MNIST}~(Table~A\ref{tab:mnist-detailed}) and 
\texttt{FashionMNIST}~(Table~A\ref{tab:fashionmnist-detailed}); 
slightly worse than the baseline for~\texttt{SVHN}~(Table~A\ref{tab:svhn-detailed}) and 
\texttt{EuroSAT}~(Table~A\ref{tab:eurosat-detailed});  
roughly on par with the baseline for~\texttt{CIFAR}~(Table~A\ref{tab:svhn-detailed}).  
All overfitted models from the Exp.~A procedure have a positive \(\tau\)-ranking 
between~\gls{icb} and the three~\gls{ge} types for all datasets 
(Table~\ref{tab:100-train}). Thus, \gls{icb} outperforms the training 
accuracy baseline ($\tau=0$) here.

For Exp.~B, there was considerable variance in \(\tau\)-rankings among the 
\(45\) binary classification tasks for each dataset. 
Although the median ranking was positive for all datasets, 
the baseline achieves a higher median ranking than ICB for 
all three error types (Clean, \gls{awgn}, 
\gls{fgsm})~(Fig.~A\ref{fig:ranking_gen_err_steady_state}).

\biblio


\section{Discussion}

Our results show that the~\gls{icb} serves as a non-vacuous generalization bound, 
which we verified in the case of infinite-width networks. Furthermore, we
performed a broader evaluation than is typically considered for theoretical GE bounds:
i) We attempted to falsify~\gls{icb} by evaluating it throughout training, rather than 
at a specific number of epochs or training loss value. 
ii) We varied the number of training samples and classification labels, 
compared to a static train/test split. iii) We considered robust~\gls{ge} in addition 
to standard~\gls{ge}. iv) Our experiments were performed on five datasets.

\gls{icb} was consistently satisfied at the expected \(95\%\) rate for models with at 
least \(70-80\%\) test accuracy, which is encouraging since accurate models 
are more likely to be deployed in practice. 
It is, however, more helpful to threshold by training accuracy to 
establish a regime in which~\gls{icb} always works well, since one does not assume 
access to held-out data when evaluating generalization bounds. 
The relationship between training accuracy and the percentage of models 
satisfying~\gls{icb} was unfortunately weaker, despite being nearly \(100\%\) 
for overfitted models.
Nonetheless,~\gls{icb} was satisfied at a high rate of at least \(92\%\) of the time 
for three of five datasets (\texttt{MNIST}, \texttt{FashionMNIST}, and~\texttt{EuroSAT}) 
without excluding any models by accuracy, and the training label randomization test was 
sensitive to tasks where ICB wasn't satisfied.

Compared to a simple training accuracy baseline,~\gls{icb} performed well at ranking~\gls{ge} 
when the classification task was allowed to vary (e.g., grouping errors for 1 vs.~2 classification 
with those for a 2 vs.~3 task), or when the training
accuracy was fixed at \(100\%\). \gls{icb}, however, did not always outperform the
training accuracy baseline for specific tasks and when~\glspl{ge} took a large range. 
However, a limited error ranking ability is not necessarily disqualifying for a generalization 
bound. It is unclear to what extent a generalization bound~\emph{ought} to be able to 
rank~\glspl{ge}, given that it is by definition merely an upper bound on the error. 
For example,~\glspl{ge} of \(1\%\) and \(29\%\) are both compatible with a bound of 
\(30\%\), which would contribute to a poor ranking in terms of Kendall's \(\tau\). 
When varying one metaparameter at a time---in particular the diagonal regularization 
coefficient---a strong monotonic relationship is observed between~\gls{icb} and robust 
errors~\gls{awgn} and~\gls{fgsm}.

\paragraph{Relevance to deep learning}
One should use caution before extrapolating our conclusions based on infinite-width 
networks to finite-width \glspl{dnn}. The ability of infinite-width networks to 
approximate their finite-width counterparts is reduced with increasing training 
samples~\citep{lee2019wide}, regularization~\citep{lee2020finitea}, and depth~\citep{li2021future}.
Nevertheless, the infinite-width framework has allowed us to demonstrate the 
practical relevance of the~\gls{icb} for an exciting family of models as a first step. 
It has been argued that understanding generalization for shallow kernel learning models is 
essential to understanding generalization behaviour of deep networks.
Kernel learning and deep learning share the ability to exactly fit their training sets yet 
still generalize well, a phenomenon that other bounds fail to explain~\citep{belkin2018understand}.
We leave the study of~\gls{icb} in the context of finite-width~\glspl{dnn} to future work, 
which may require alternative MI estimation techniques.

\biblio


\section{Related Work}

\paragraph{Kernel-regression generalization error}
\cite{canatar2021spectral} derived an analytical expression for the 
generalization~\gls{mse} of kernel regression models using a replica 
method from statistical mechanics. Their predictions show excellent 
agreement with the empirical~\gls{ge} of~\gls{ntk} models on~\texttt{MNIST} 
and~\texttt{CIFAR} datasets as a function of the training sample size. 
Furthermore, their method is sensitive to differences in difficulty 
between similar classification tasks, e.g., showing that~\texttt{MNIST} ``0 vs 1'' 
digit classification is easier to learn than ``8 vs 9''. Their method has also
been extended to predict out-of-distribution generalization 
error~\citep{canatar2021outofdistribution}.

An alternative method is the~\gls{loo} error  estimator~\citep{lachenbruch1967almost}. 
\gls{loo} is generally impractical for deep learning due to the computational 
effort required to train $n$ \glspl{dnn} from scratch on $n$ different training sets. 
However, \cite{bachmann2021generalization} proposed a closed-form expression 
for the~\gls{loo} estimator for a single kernel regression model trained once on the 
full training set. 
Their estimator shows excellent agreement with test~\gls{mse} and accuracy
for a five-layer ReLU NTK model trained on~\texttt{MNIST} and~\texttt{CIFAR}. 
While~\citeauthor{bachmann2021generalization} averaged results over 
five training sets of size \(500-20000\), we only draw a single training 
set of \(250-2000\) samples for each set of metaparameters. Our choice was made 
to reflect a practical ``small data'' scenario, where~\gls{ge} 
has to be bounded using a modest set of labeled data. As a result, however, our~\gls{ge}
and ICB estimates have greater variance than those of~\citeauthor{bachmann2021generalization}.

In this work, we used the infinite-width limit of neural networks for convenience 
and as a first step to assess the efficacy of~\gls{icb}; we did not set out to find 
optimal generalization bounds for kernel regression. An advantage of~\gls{icb} is that 
it only requires access to \(I(X; Z)\)---a black-box statistic applicable to a wide 
variety of models beyond kernel regression. Therefore, \gls{icb} may become increasingly 
relevant for deep learning using MI estimators with different assumptions, e.g., 
with distributional constraints on weight matrices~\citep{gabrie2018entropy} or 
log-Gaussian corrections for infinite depth-and-width limits~\citep{li2021future}.

\paragraph{Generalization bounds for deep learning}
\cite{dziugaite2017computing} develop a PAC-Bayes~\gls{ge} bound and evaluated it 
on a~\texttt{MNIST} binary classification task (classes $\{0, \dots, 4\}$ vs.~$\{5, \dots, 9\}$) 
using the complete training set ($N_\text{train} = 55k$) 
and a ReLU fully-connected~\gls{nn} with 2-3 layers. 
Although their bound was non-vacuous (\(\approx 20\%\)), it was several times 
larger than the error estimated on held-out data (\(<1\%\)).
A direct comparison with our work is difficult, as we did not use finite-width~\glspl{dnn}. 
We showed that the~\gls{icb} yields a smaller (\(\approx 10\%)\) bound from less than 
\(2000\) samples for~\texttt{MNIST} binary classification tasks. 
The PAC-Bayes bound was also evaluated at a single point during training at 20 
epochs, whereas we evaluated ICB from initialization to convergence to check 
for violations and tightness of the bound.

\cite{zhou2019nonvacuousa} proposed a PAC-Bayes 
generalization bound based on the compressed size of a~\gls{dnn} after 
pruning and quantization. They obtain a~\gls{ge} bound of \(46\%\) for MNIST 
and \(96-98\%\) for ImageNet. The measure of compression used by~\cite{zhou2019nonvacuousa} 
should not be confused with input compression in terms of~\gls{mi} in this study. The bounds 
of~\cite{dziugaite2017computing} and~\cite{zhou2019nonvacuousa} concern model 
complexity, whereas~\gls{icb} is concerned with input data compression by the hidden 
layers. Both~\citeauthor{dziugaite2017computing} and~\citeauthor{zhou2019nonvacuousa} retrained 
models from scratch to optimize their bound for best results, whereas 
we did not modify any aspect of the standard gradient descent training 
procedure or~\gls{ntk} architecture to be more aligned with~\gls{icb}.

\paragraph{Input compression and deep learning}
\cite{saxe2018information} observed a lack of compression in 
ReLU networks and argued that compression 
must be unrelated to generalization in~\glspl{dnn}, 
since it is known that ReLU networks generalize well. 
However, their binning procedure based on~\citep{paninski2003estimation} 
involves metaparameters that influence entropy and~\gls{mi} estimation. 
Other works reported input compression in linear regression~\citep{chechik2005informationa} 
and finite-width ReLU~\glspl{dnn} using adaptive binning estimators~\citep{chelombiev2019adaptive}.
We use~\gls{mi} bounds free from such metaparameters and observe input 
compression regardless of the nonlinearity type, consistent 
with~\cite{shwartz-ziv2020information}.

The construction of invertible~\glspl{dnn}~\citep{jacobsen2018irevnet} that generalize well 
has raised questions about a possible connection between compression and generalization 
of~\glspl{dnn}. However, to the best of our knowledge there have been few attempts to 
accurately measure information-theoretic compression in deterministic invertible 
networks.\footnote{\cite{ardizzone2020exact} formulate an Information Bottleneck loss 
for invertible networks with additive noise.} 
Although~\gls{mi} diverges for deterministic networks~\citep{goldfeld2019estimating}, 
input compression may be observed in principle by substituting entropy
for~\gls{mi}~\citep{strouse2016deterministica}. 
This noise-free limit has also been studied in the context of information
maximization~\citep{bell1995informationmaximization}.
We are excited about future work that may shed light on input compression phenomena 
and the challenging case of deep, deterministic, finite-width~\glspl{nn}.

\biblio


\section{Conclusion}

We assessed the~\gls{icb} along three axes of performance: tightness of the bound 
(e.g., vacuous or non-vacuous), percentage of trials satisfying the bound, and 
ranking generalization errors. 
Our empirical results show that input compression serves as a simple and effective 
generalization bound, complementing previous theoretical evidence.
Additionally, \gls{icb} can help pinpoint interesting failures of 
robust generalization that go undetected by standard generalization metrics.
An important consequence of the~\gls{icb} with respect to~\gls{nas} is 
that~\emph{bigger is not necessarily better}, at least in terms of the information complexity of 
infinite-width networks. Equally if not more important than the architecture are the 
metaparameters and duration of model training, all of which influence input compression.
Consistent with Occam's razor, less information complexity---or more input
compression---yields more performant models, reducing the upper bound on generalization error. 
We conclude that input compression, which is data-centric, is a more effective complexity 
metric than model-centric proxies like the number of parameters or depth. 

\biblio

\subsubsection*{Acknowledgments}
This research was developed with funding from the Defense Advanced Research Projects Agency (DARPA).
The views, opinions and/or findings expressed are those of the author and should not be interpreted as
representing the official views or policies of the Department of Defense or the U.S. Government.
Graham W.~Taylor and Angus Galloway also acknowledge support from CIFAR and the Canada Foundation
for Innovation. Angus Galloway also acknowledges supervision by Medhat Moussa.
Resources used in preparing this research were provided, in part, by the Province of Ontario,
the Government of Canada through CIFAR, and companies sponsoring the Vector
Institute: \url{http://www.vectorinstitute.ai/#partners}.
Anna Golubeva is supported by the National Science Foundation under Cooperative Agreement
PHY-2019786 (The NSF AI Institute for Artificial Intelligence and Fundamental Interactions,
http://iaifi.org/). Mihai Nica is supported by an NSERC Discovery Grant.

\bibliography{tmlr_bibliography}

\bibliographystyle{tmlr}

\clearpage


\appendix

\section{Appendix}


\subsection{Lower bound on~\gls{mi}}
\label{sec:sub-appendix-lower-bound-mi}

We may lower bound \(I(X; Z)\) using a bound of similar form as~\eqref{eq:ub}
based on a batch of \(N\) samples:

\begin{equation} \label{eq:lb}
I(X; Z) \geq \E \left[ \frac{1}{N} \sum_{i=1}^{N} \log
\frac{p(z_i | x_i)}{\frac{1}{N} \sum_j p(z_i | x_j)} \right] = I_{\text{LB}} \,,
\end{equation}
where the expectation is taken over \(N\) independent samples from the joint distribution \(\prod_j p(x_j, z_j)\). The main difference between this bound and~\eqref{eq:lb} is the inclusion of \(p(z_i | x_i) \) in the denominator.

\subsection{Illustrative example and filtering~\gls{mi}}
\label{sec:sub-appendix-mi-example}

\begin{figure}

\centering
\includegraphics[width=\textwidth]{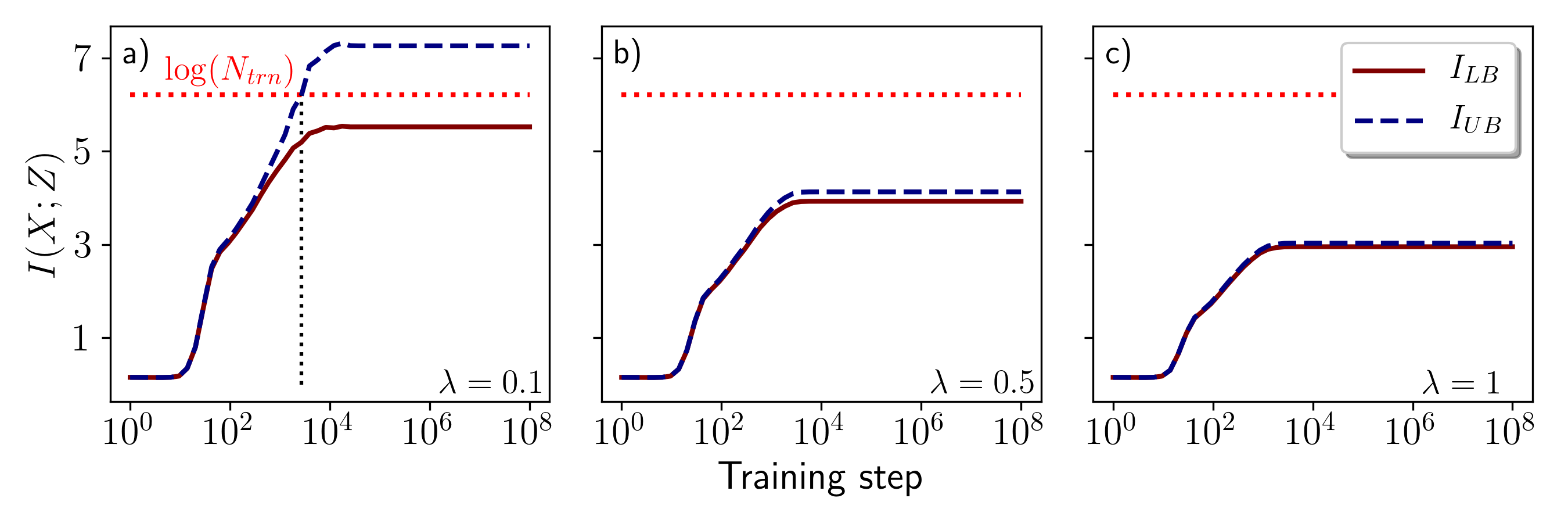}
\caption{We plot \(I(X; Z)\) upper (\ref{eq:ub}) and lower (\ref{eq:lb}) bounds
corresponding to the illustrative~\texttt{EuroSAT} example (Figure~\ref{fig:euro-teaser}).
Increasing the regularization to \(\lambda=0.5\) in b) and \(\lambda=1.0\) in c)
reduces~\gls{mi} below \(\log(N_\text{trn})\). Samples to the right of the vertical line in
a) where \(I_{UB}\) crosses \( \log(N_\text{trn}) \) are discarded for the main analyses. NB:
We use natural units (``Nats'' or ``Shannons'') for \(I(X; Z)\) here, but we convert to bits
when evaluating the~\gls{icb}.}
\label{fig:euro-teaser-mi}
\end{figure}

We empirically verified that~\eqref{eq:lb} and~\eqref{eq:ub} yield
similar results when \(I_\text{UB} < \log(N_\text{trn})\) (Fig.~A\ref{fig:euro-teaser-mi}).

\subsection{Bounding generalization throughout training}
\label{sec:sub-appendix-through-training}

\begin{figure}
\centering
\includegraphics[width=\textwidth]{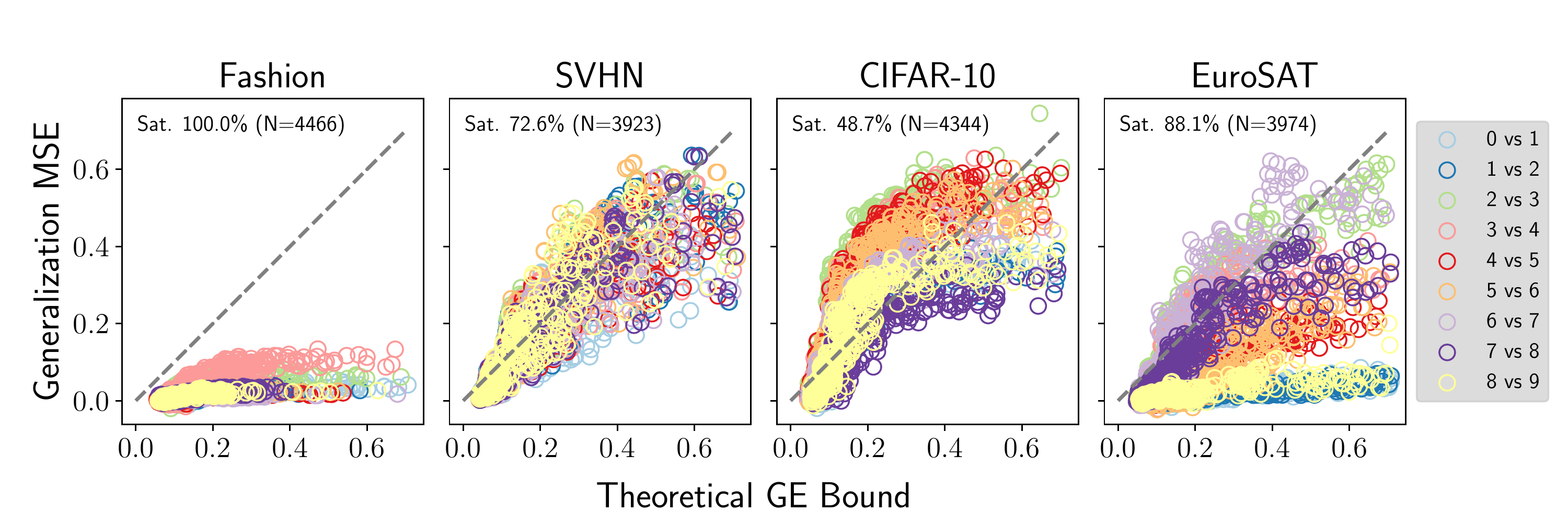}
\caption{\gls{icb} is plotted versus~\gls{ge} for~\texttt{FashionMNIST},
\texttt{SVHN}, \texttt{CIFAR-10}, and \texttt{EuroSAT} datasets.
The~\gls{icb} satisfaction rate is annotated in the top left corner of each
plot with format ``ICB \% (N)''.
Each binary classification task is assigned a unique colour to highlight
inter-task differences in~\gls{icb} satisfaction rate.
See Figure~\ref{fig:bound_ge_acc_throughout_training} of \S\ref{sec:sub-res-bounding-ge}
for the corresponding Figure with~\gls{ge} expressed in terms of classification error
rather than MSE. NB: Results for~\texttt{MNIST} omitted from Figure as they were
similar to~\texttt{FashionMNIST}.}
\label{fig:bound_ge_mse_throughout_training}
\end{figure}

\paragraph{Loss function}
We considered~\gls{ge} in terms of~\gls{mse} in addition to classification error (Fig.~A\ref{fig:bound_ge_mse_throughout_training}).
This change results in no difference in the overall~\gls{icb}
Sat.~for~\texttt{FashionMNIST}, an improvement for~\texttt{SVHN} from
\(44.5\%\) to \(72.6\%\), and a small decrease for~\texttt{CIFAR-10} from
to \(59.3\%\) to \(48.7\%\) as well as for~\texttt{EuroSAT} from \(93.6\%\)
to \(88.1\%\).

\paragraph{Activation function}
Overall, \texttt{ReLU} networks satisfied~\gls{icb} more frequently than~\texttt{Erf}
networks (Table~A\ref{tab:ten-choose-two-bounding-error-steady-state}).


The following caption applies to Tables~A\ref{tab:mnist-detailed}-\ref{tab:eurosat-detailed}. \par

\begin{mdframed}
Kendall's $\tau$ ranking for three generalization error types: Clean, \gls{awgn}
and \gls{fgsm} by training accuracy ``Train (baseline)'' and ICB are presented.
One hundred random seeds are used to draw different metaparameters uniform random
for each task, for which models are evaluated five (5) times each during training
resulting in a maximum of 500 samples per task. The number of valid samples out of 500,
i.e., those with $I_\text{UB}(X; Z) \leq \log(N_\text{train})$ is indicated in
the $N_\text{valid}$ column. ICB \% indicates the percentage of samples that satisfy
the~\gls{icb}, i.e., \texttt{Clean generalization error} $ \leq$ \texttt{ICB}.
Entries in the ``Row average'' row are obtained by simply averaging across the nine (9) tasks.
Kendall's $\tau$ values for the ``Overall'' row may differ substantially from ``Row average''
as this corresponds to aggregating all raw data points and considering them as one task
before calculating $\tau$\@.~\textbf{Bold} is used to denote whether the baseline or ICB
achieve a better ranking of generalization errors.
\end{mdframed}


\begin{figure}
\centering
\begin{subfigure}{\textwidth}
\centering
\includegraphics[width=.8\textwidth]{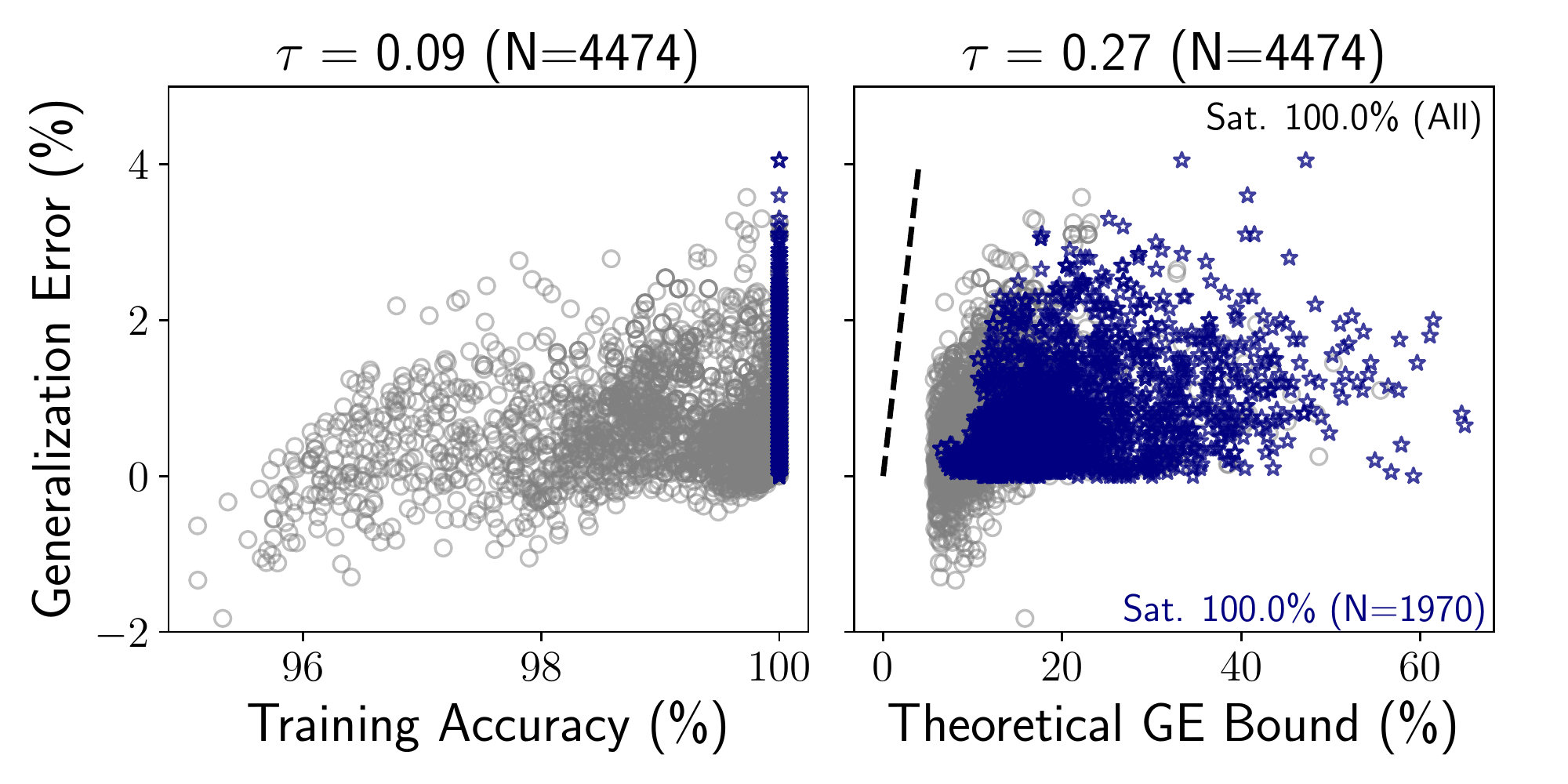}
\caption{}\label{fig:mnist-v5}
\end{subfigure}
\begin{subfigure}{\textwidth}
\centering
\includegraphics[width=.8\textwidth]{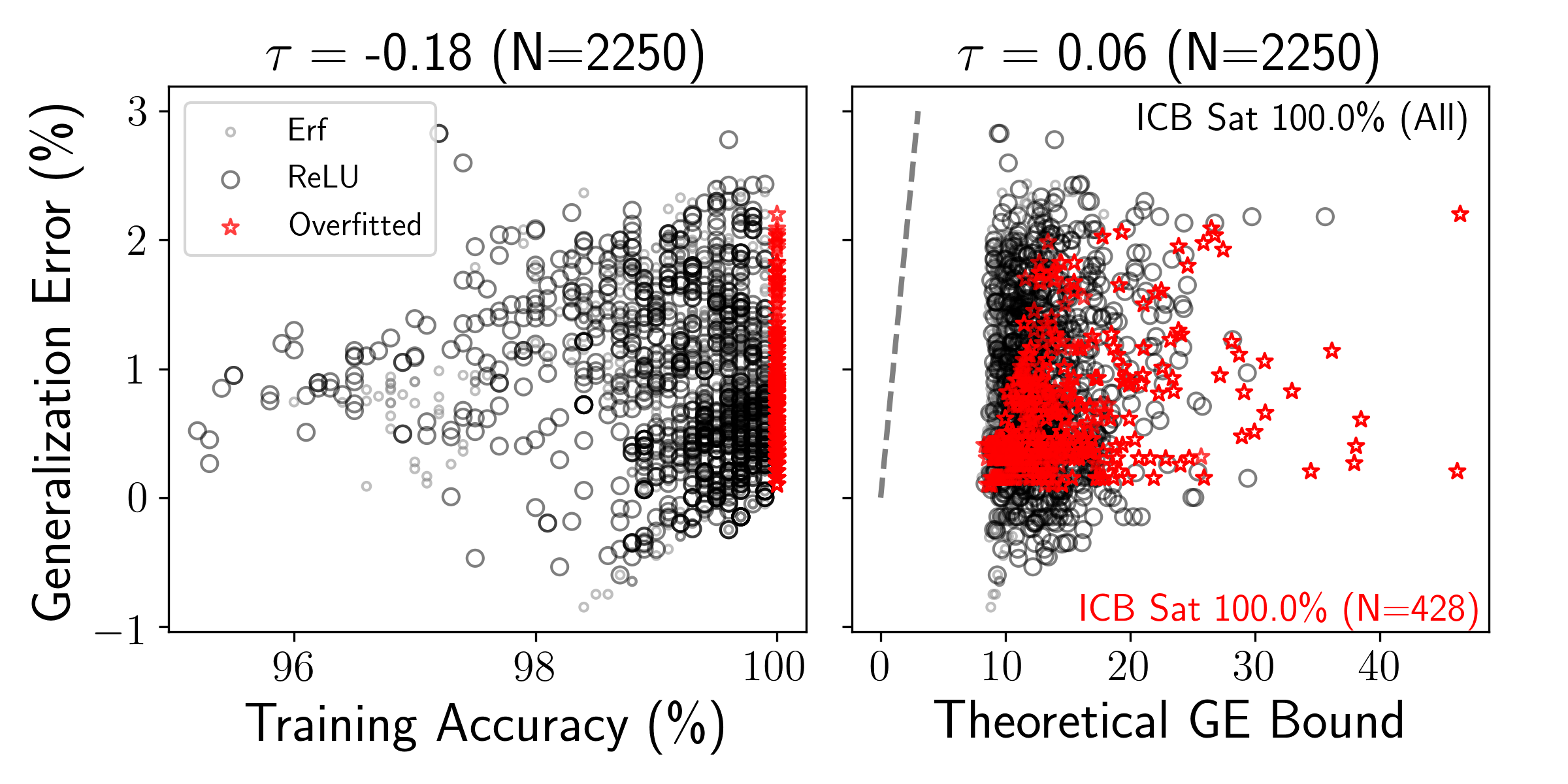}
\caption{}\label{fig:mnist-v2-steady}
\end{subfigure}
\caption{\subref{fig:mnist-v5}) MNIST models evaluated throughout training,
\subref{fig:mnist-v2-steady}) models evaluated at steady state \(t=\infty\).}
\label{fig:mnist-scatter}
\end{figure}

\begin{table}[]
\centering
\caption{Kendall's \(\tau\) ranking for~\texttt{MNIST}. See complete caption in~\S\ref{sec:sub-appendix-through-training}.}
\label{tab:mnist-detailed}
\begin{tabular}{ccccccccc}
\toprule
\multicolumn{3}{c}{Details} & \multicolumn{3}{c}{Train (baseline)} & \multicolumn{3}{c}{ICB} \\ \midrule
Task & $N_\text{valid}$ & ICB \% & Clean & AWGN & FGSM & Clean & AWGN & FGSM \\ \midrule
0 vs. 1 & 499 & 100\% & 0.36 & 0.37 & 0.35 & 0.10 & 0.09 & 0.10 \\
1 vs. 2 & 498 & 100\% & 0.39 & 0.65 & 0.33 & 0.34 & 0.52 & 0.32 \\
2 vs. 3 & 497 & 100\% & 0.42 & 0.57 & 0.46 & 0.42 & 0.55 & 0.46 \\
3 vs. 4 & 500 & 100\% & 0.27 & 0.32 & 0.24 & 0.22 & 0.29 & 0.23 \\
4 vs. 5 & 493 & 100\% & 0.19 & 0.31 & 0.14 & 0.19 & 0.32 & 0.17 \\
5 vs. 6 & 497 & 100\% & 0.49 & 0.59 & 0.51 & 0.44 & 0.53 & 0.46 \\
6 vs. 7 & 498 & 100\% & 0.29 & 0.27 & 0.23 & 0.17 & 0.23 & 0.17 \\
7 vs. 8 & 498 & 100\% & 0.35 & 0.44 & 0.32 & 0.38 & 0.46 & 0.38 \\
8 vs. 9 & 494 & 100\% & 0.42 & 0.56 & 0.44 & 0.38 & 0.52 & 0.42 \\ \midrule
\multicolumn{2}{r}{Row average} & \multirow{2}{*}{100\%} & \textbf{0.35} & \textbf{0.45} & \textbf{0.34} & 0.29 & 0.39 & 0.30 \\
\multicolumn{2}{r}{Overall} &  & 0.09 & 0.12 & 0.03 & \textbf{0.27} & \textbf{0.31} & \textbf{0.24} \\ \bottomrule
\end{tabular}
\end{table}


\begin{figure}
\centering
\begin{subfigure}{\textwidth}
\centering
\includegraphics[width=.8\textwidth]{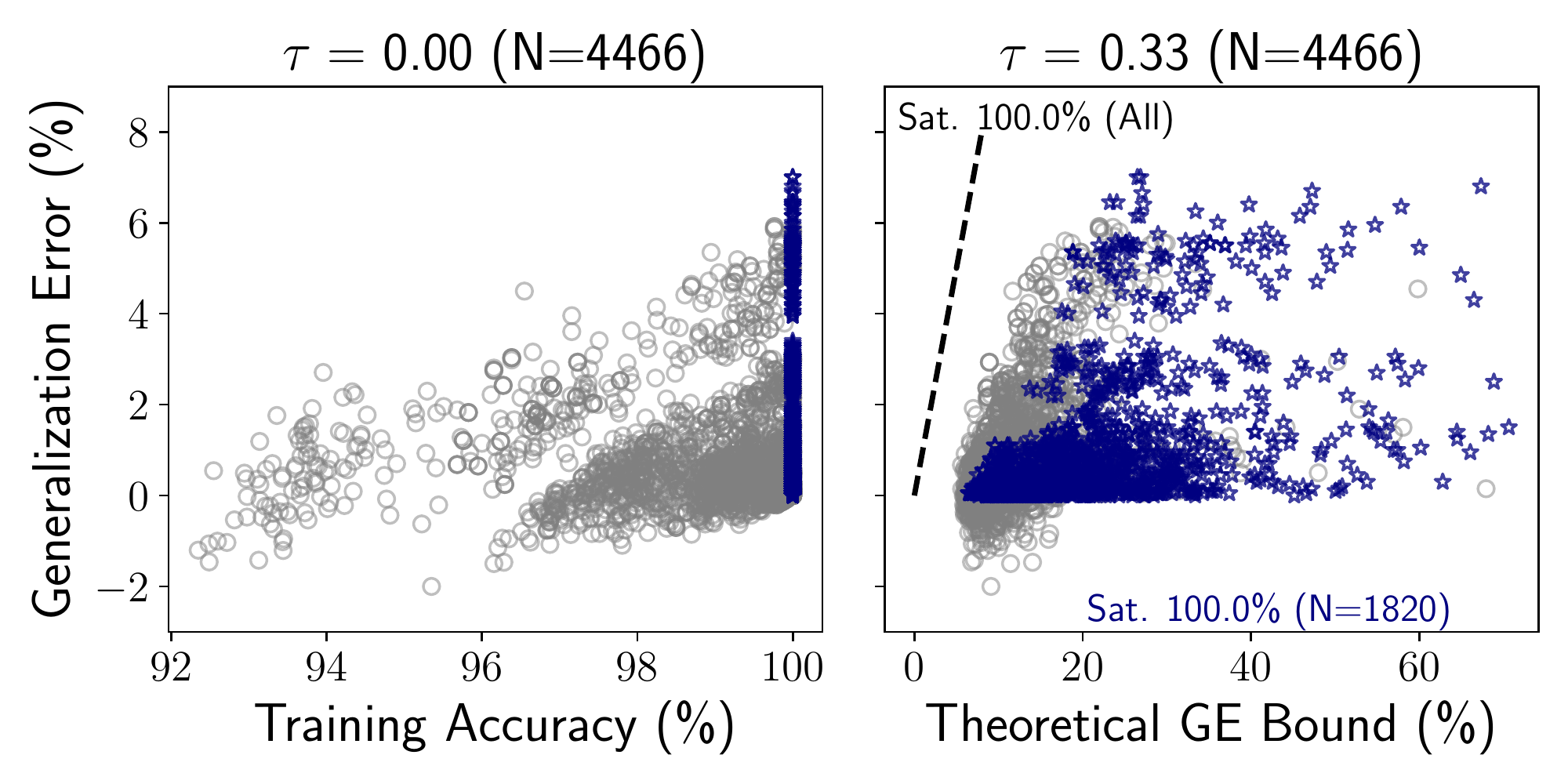}
\caption{}\label{fig:fashion_mnist-v5}
\end{subfigure}
\begin{subfigure}{\textwidth}
\centering
\includegraphics[width=.8\textwidth]{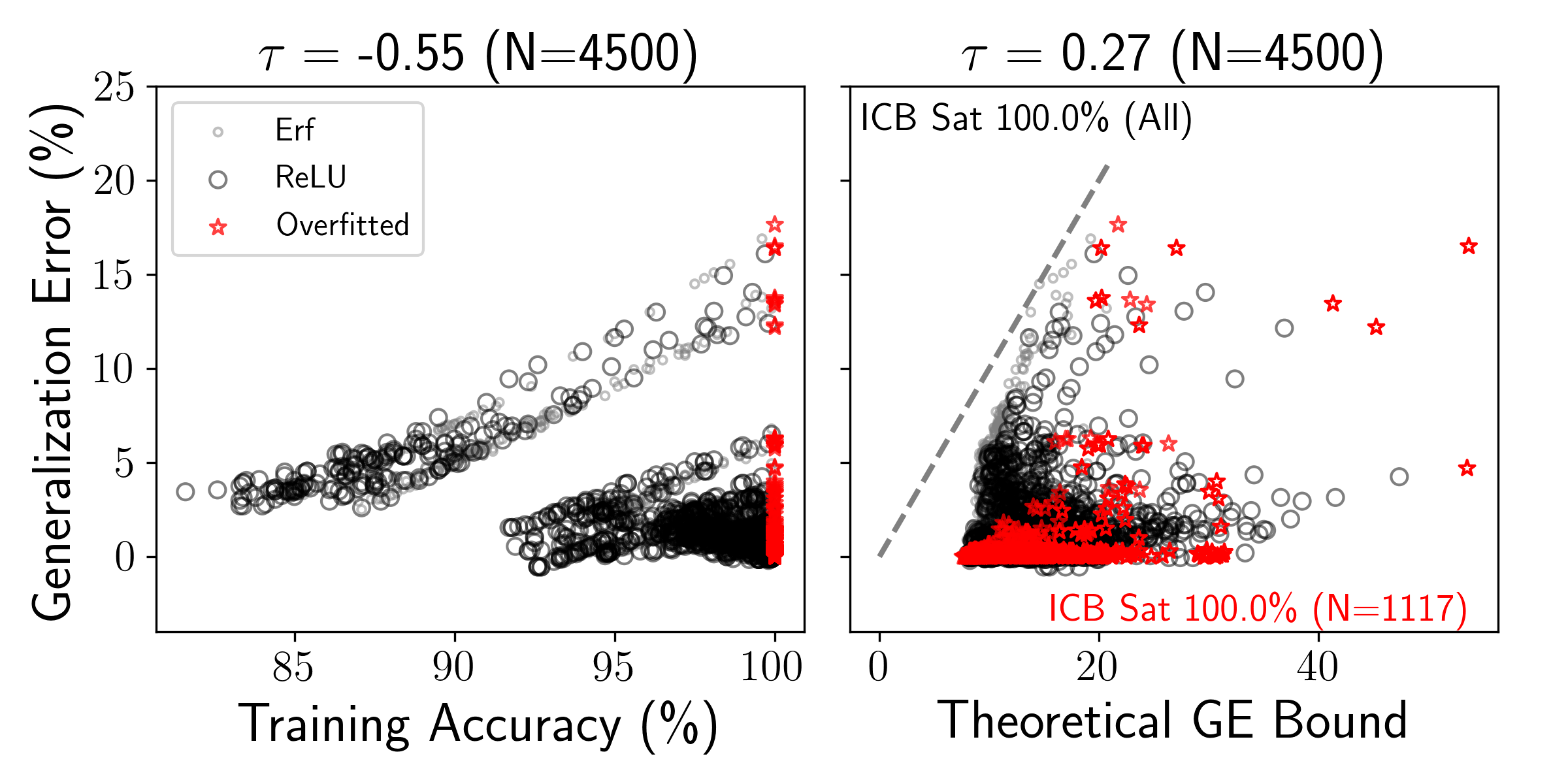}
\caption{}\label{fig:fashion-v2-steady}
\end{subfigure}
\caption{FashionMNIST models evaluated \subref{fig:fashion_mnist-v5}) throughout
training and \subref{fig:fashion-v2-steady}) at steady state \(t=\infty\).}
\label{fig:fashion-scatter}
\end{figure}

\begin{table}[]
\centering
\caption{Kendall's \(\tau\) ranking for~\texttt{FashionMNIST}. See complete caption in~\S\ref{sec:sub-appendix-through-training}.}
\label{tab:fashionmnist-detailed}
\begin{tabular}{ccccccccc}
\toprule
\multicolumn{3}{c}{Details} & \multicolumn{3}{c}{Train (baseline)} & \multicolumn{3}{c}{ICB} \\ \midrule
Task & $N_\text{valid}$ & ICB \% & Clean & AWGN & FGSM & Clean & AWGN & FGSM \\ \midrule
0 vs. 1 & 250 & 100\% & 0.59 & 0.67 & 0.66 & 0.46 & 0.57 & 0.54 \\
1 vs. 2 & 248 & 100\% & 0.48 & 0.59 & 0.55 & 0.42 & 0.50 & 0.46 \\
2 vs. 3 & 247 & 100\% & 0.76 & 0.80 & 0.80 & 0.57 & 0.60 & 0.61 \\
3 vs. 4 & 243 & 100\% & 0.78 & 0.82 & 0.84 & 0.67 & 0.66 & 0.66 \\
4 vs. 5 & 250 & 100\% & 0.04 & -0.02 & -0.11 & 0.08 & -0.02 & 0.00 \\
5 vs. 6 & 249 & 100\% & 0.18 & 0.15 & 0.04 & 0.13 & 0.17 & 0.06 \\
6 vs. 7 & 250 & 100\% & 0.35 & 0.26 & 0.23 & 0.17 & 0.16 & 0.12 \\
7 vs. 8 & 250 & 100\% & 0.37 & 0.52 & 0.31 & 0.35 & 0.46 & 0.35 \\
8 vs. 9 & 250 & 100\% & 0.40 & 0.29 & 0.38 & 0.20 & 0.11 & 0.20 \\ \midrule
\multicolumn{2}{r}{Row average} & \multirow{2}{*}{100\%} & \textbf{0.44} & \textbf{0.45} & \textbf{0.41} & 0.34 & 0.35 & 0.34 \\
\multicolumn{2}{r}{Overall} &  & -0.02 & -0.03 & -0.09 & \textbf{0.33} & \textbf{0.33} & \textbf{0.31} \\ \bottomrule
\end{tabular}
\end{table}


\begin{figure}
\centering
\begin{subfigure}{\textwidth}
\centering
\includegraphics[width=.8\textwidth]{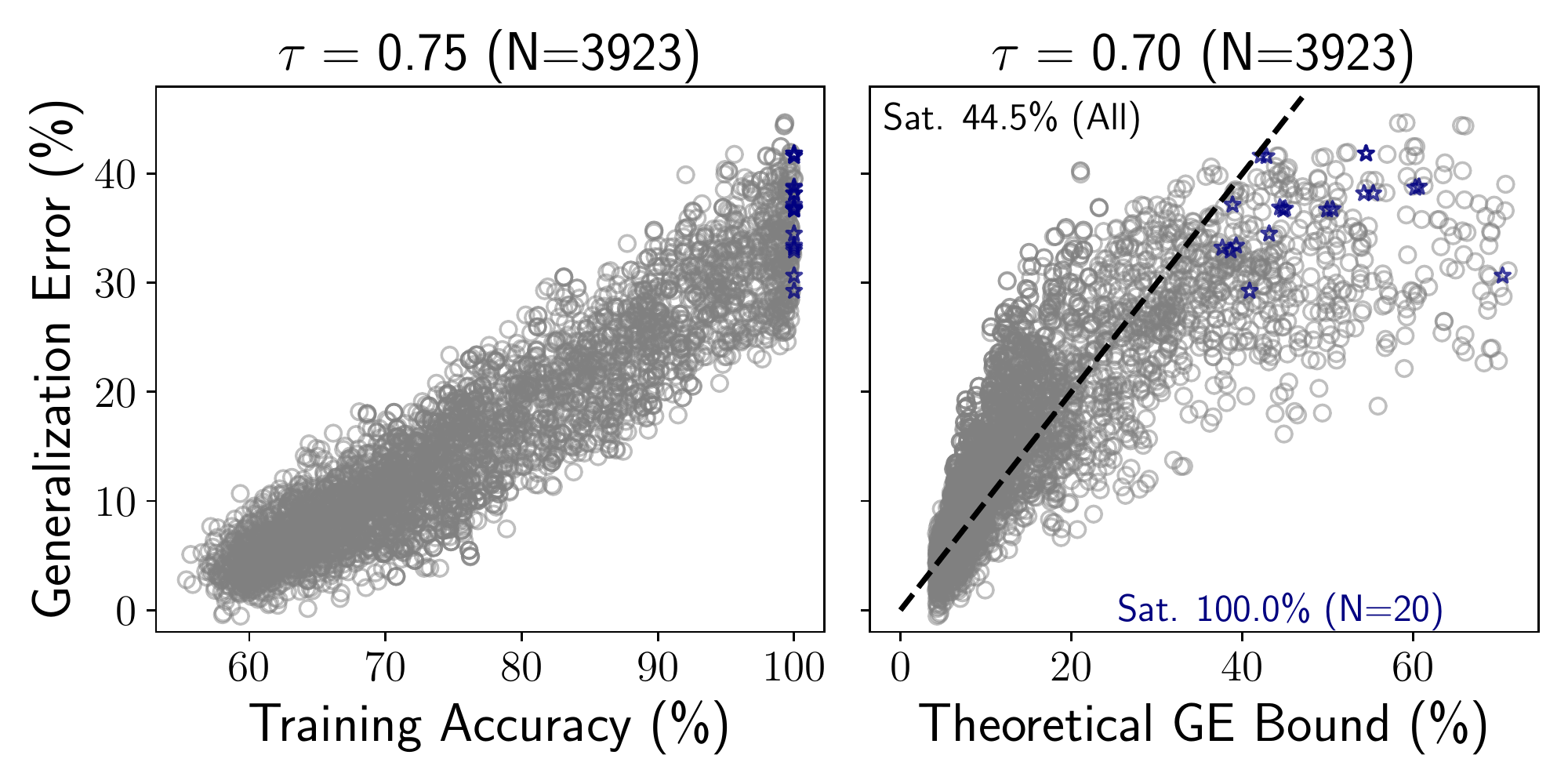}
\caption{}\label{fig:svhn-v5}
\end{subfigure}
\begin{subfigure}{\textwidth}
\centering
\includegraphics[width=.8\textwidth]{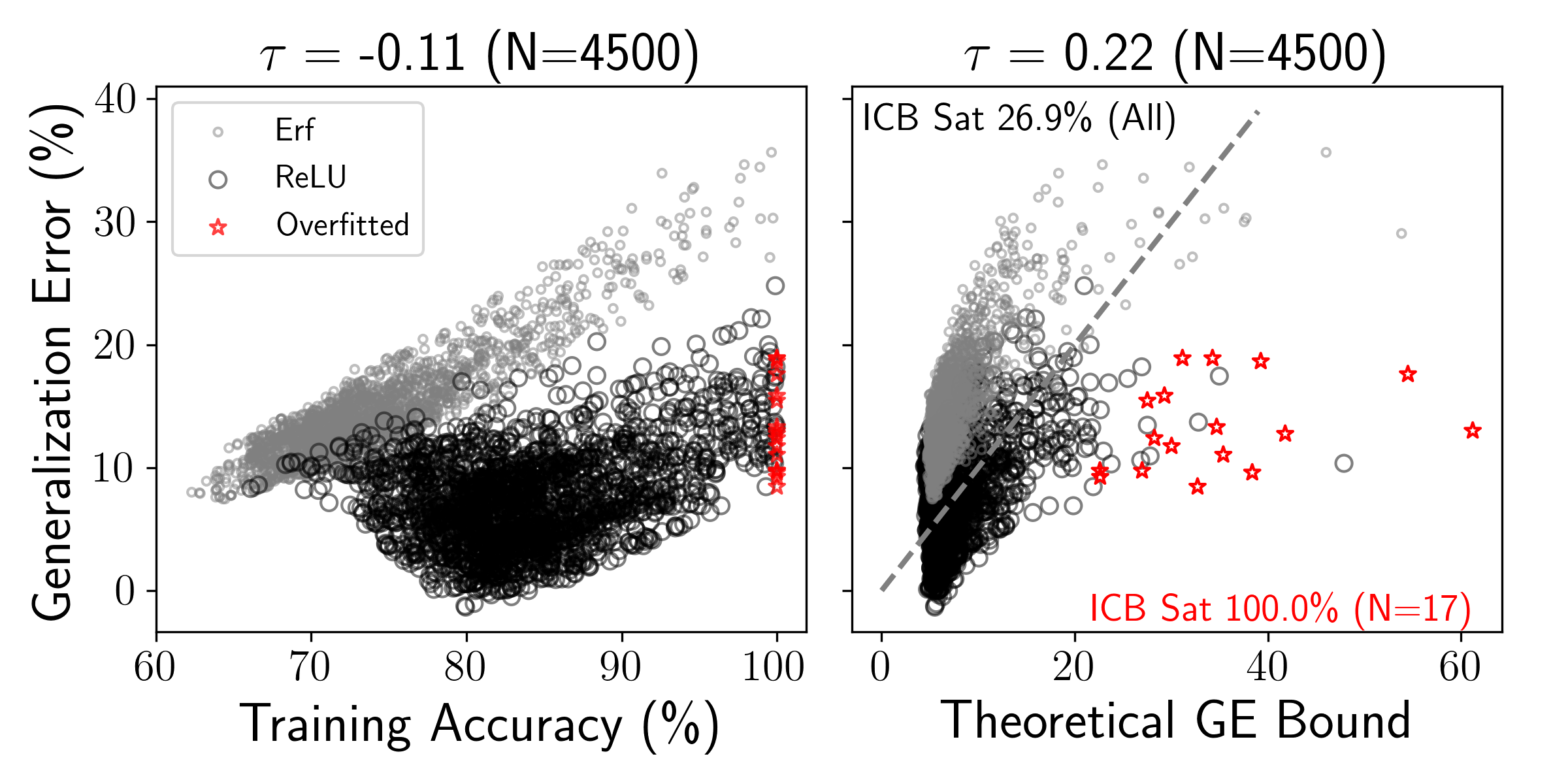}
\caption{}\label{fig:svhn-v2-steady}
\end{subfigure}
\caption{\subref{fig:svhn-v5}) SVHN models evaluated throughout
training, \subref{fig:svhn-v2-steady}) models evaluated at steady state \(t=\infty\).}
\label{fig:svhn-scatter}
\end{figure}

\begin{table}[]
\centering
\caption{Kendall's \(\tau\) ranking for~\texttt{SVHN}. See complete caption in~\S\ref{sec:sub-appendix-through-training}.}
\label{tab:svhn-detailed}
\begin{tabular}{ccccccccc}
\toprule
\multicolumn{3}{c}{Details} & \multicolumn{3}{c}{Train (baseline)} & \multicolumn{3}{c}{ICB} \\ \midrule
Task & $N_\text{valid}$ & ICB \% & Clean & AWGN & FGSM & Clean & AWGN & FGSM \\ \midrule
0 vs. 1 & 443 & 71\% & 0.75 & 0.85 & 0.80 & 0.70 & 0.74 & 0.63 \\
1 vs. 2 & 432 & 33\% & 0.81 & 0.90 & 0.84 & 0.74 & 0.73 & 0.66 \\
2 vs. 3 & 440 & 34\% & 0.82 & 0.89 & 0.83 & 0.74 & 0.73 & 0.64 \\
3 vs. 4 & 438 & 44\% & 0.82 & 0.90 & 0.83 & 0.73 & 0.75 & 0.67 \\
4 vs. 5 & 441 & 60\% & 0.81 & 0.90 & 0.83 & 0.76 & 0.75 & 0.65 \\
5 vs. 6 & 442 & 25\% & 0.86 & 0.92 & 0.86 & 0.73 & 0.72 & 0.65 \\
6 vs. 7 & 429 & 49\% & 0.79 & 0.90 & 0.83 & 0.72 & 0.71 & 0.63 \\
7 vs. 8 & 440 & 48\% & 0.78 & 0.88 & 0.83 & 0.73 & 0.73 & 0.65 \\
8 vs. 9 & 438 & 32\% & 0.84 & 0.91 & 0.82 & 0.74 & 0.73 & 0.63 \\ \midrule
\multicolumn{2}{r}{Row average} & \multirow{2}{*}{44\%} & \textbf{0.81} & \textbf{0.89} & \textbf{0.83} & 0.73 & 0.73 & 0.65 \\
\multicolumn{2}{r}{Overall} &  & \textbf{0.75} & \textbf{0.85} & \textbf{0.80} & 0.71 & 0.72 & 0.64 \\ \bottomrule
\end{tabular}
\end{table}


\begin{figure}
\centering
\begin{subfigure}{\textwidth}
\centering
\includegraphics[width=.8\textwidth]{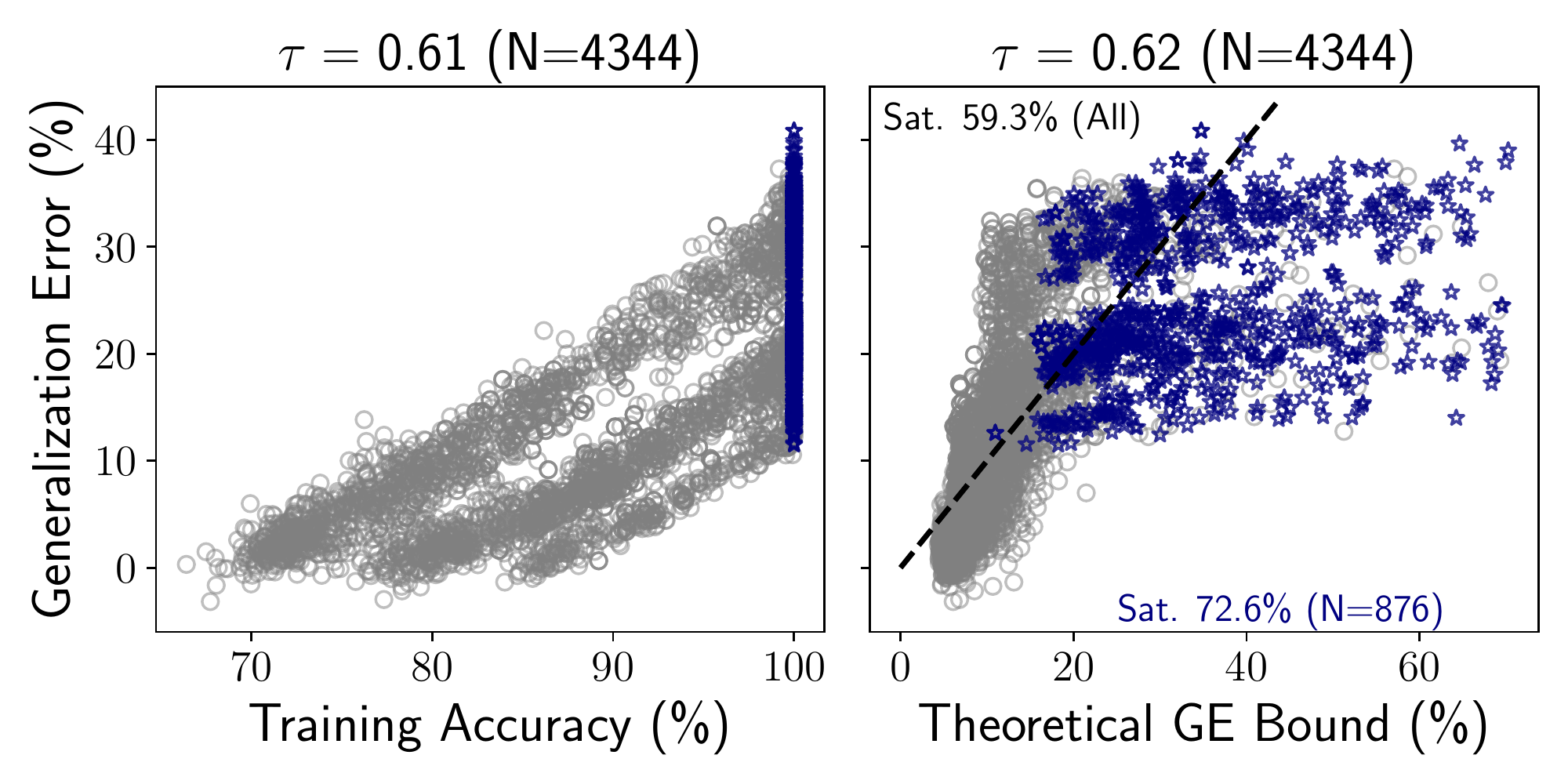}
\caption{}\label{fig:cifar-v5}
\end{subfigure}
\begin{subfigure}{\textwidth}
\centering
\includegraphics[width=.8\textwidth]{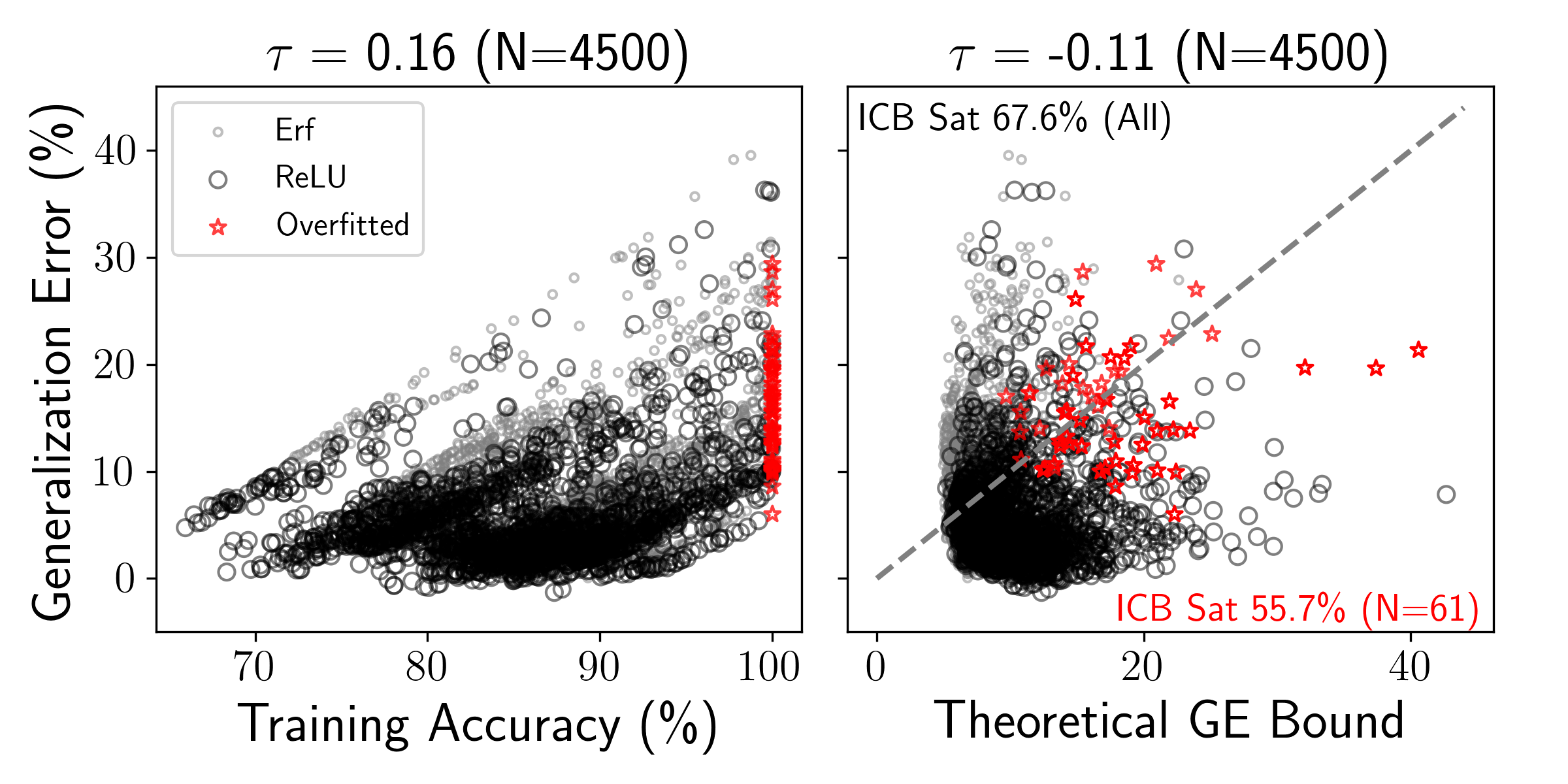}
\caption{}\label{fig:cifar-v2-steady}
\end{subfigure}
\begin{subfigure}{\textwidth}
\centering
\includegraphics[width=.8\textwidth]{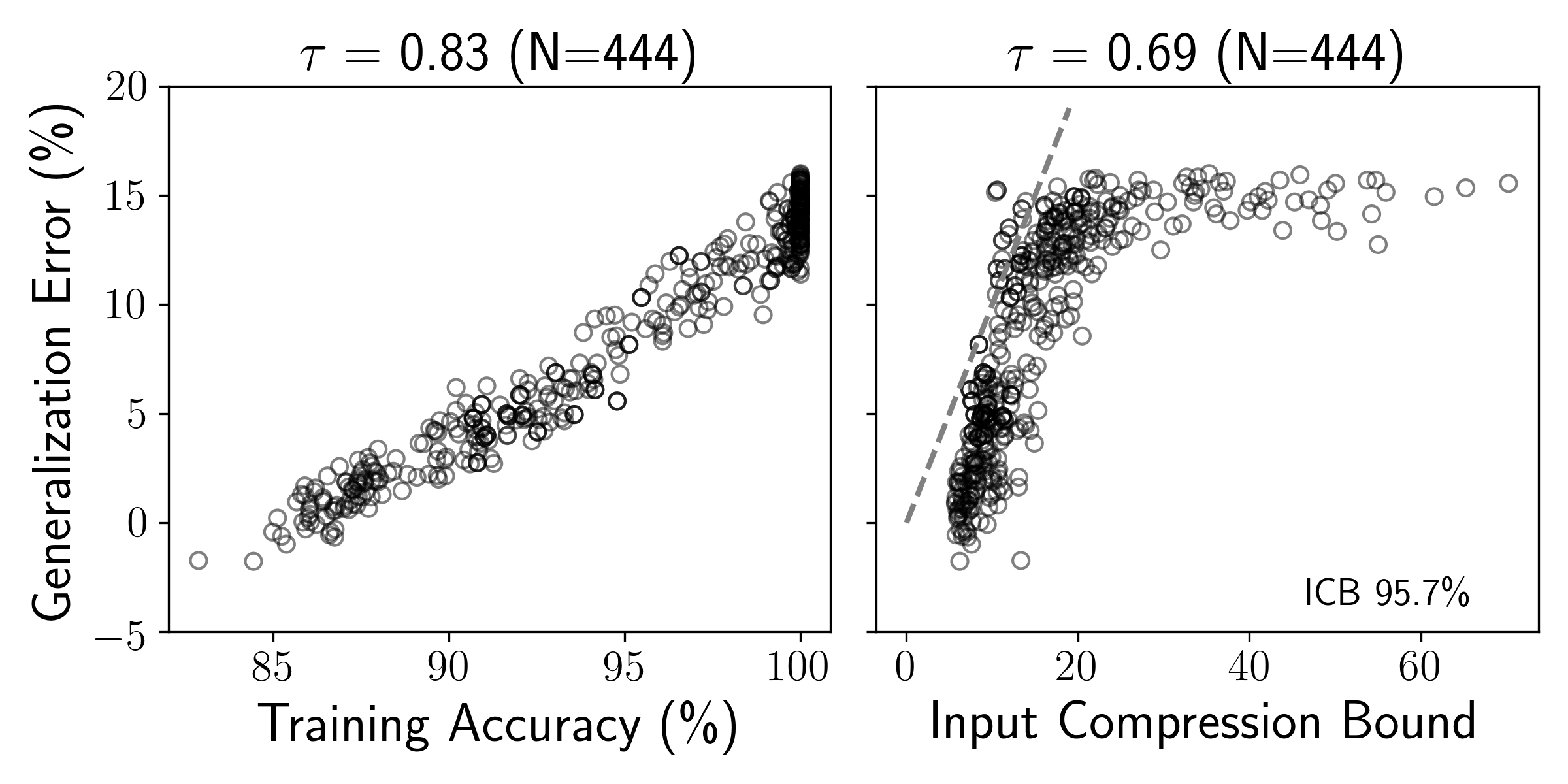}
\caption{}\label{fig:cifar-v4-test-acc-84}
\end{subfigure}
\caption{\subref{fig:cifar-v5}) CIFAR models evaluated throughout training,
\subref{fig:cifar-v2-steady}) models evaluated at steady state \(t=\infty\),
\subref{fig:cifar-v4-test-acc-84}) select models from \subref{fig:cifar-v5})
with $\geq 84 \%$ test accuracy.}
\label{fig:cifar-scatter}
\end{figure}

\begin{table}[]
\centering
\caption{Kendall's \(\tau\) ranking for~\texttt{CIFAR-10}. See complete caption in~\S\ref{sec:sub-appendix-through-training}.}
\label{tab:cifar-detailed}
\begin{tabular}{ccccccccc}
\toprule
\multicolumn{3}{c}{Details} & \multicolumn{3}{c}{Train (baseline)} & \multicolumn{3}{c}{ICB} \\ \midrule
Task & $N_\text{valid}$ & ICB \% & Clean & AWGN & FGSM & Clean & AWGN & FGSM \\ \midrule
0 vs. 1 & 482 & 74\% & 0.85 & 0.88 & 0.87 & 0.72 & 0.73 & 0.69 \\
1 vs. 2 & 489 & 79\% & 0.84 & 0.87 & 0.87 & 0.75 & 0.75 & 0.70 \\
2 vs. 3 & 487 & 29\% & 0.88 & 0.90 & 0.86 & 0.77 & 0.76 & 0.68 \\
3 vs. 4 & 480 & 35\% & 0.86 & 0.88 & 0.85 & 0.76 & 0.76 & 0.69 \\
4 vs. 5 & 472 & 39\% & 0.89 & 0.90 & 0.87 & 0.76 & 0.75 & 0.68 \\
5 vs. 6 & 481 & 41\% & 0.86 & 0.89 & 0.86 & 0.76 & 0.75 & 0.67 \\
6 vs. 7 & 487 & 68\% & 0.82 & 0.85 & 0.84 & 0.78 & 0.77 & 0.70 \\
7 vs. 8 & 488 & 97\% & 0.82 & 0.86 & 0.84 & 0.71 & 0.71 & 0.66 \\
8 vs. 9 & 486 & 74\% & 0.85 & 0.88 & 0.87 & 0.75 & 0.74 & 0.71 \\ \midrule
\multicolumn{2}{r}{Row average} & \multirow{2}{*}{59\%} & \textbf{0.85} & \textbf{0.88} & \textbf{0.86} & 0.75 & 0.75 & 0.69 \\
\multicolumn{2}{r}{Overall} &  & 0.61 & 0.65 & \textbf{0.64} & \textbf{0.62} & 0.65 & 0.62 \\ \bottomrule
\end{tabular}
\end{table}


\begin{figure}
\centering
\begin{subfigure}{\textwidth}
\centering
\includegraphics[width=.8\textwidth]{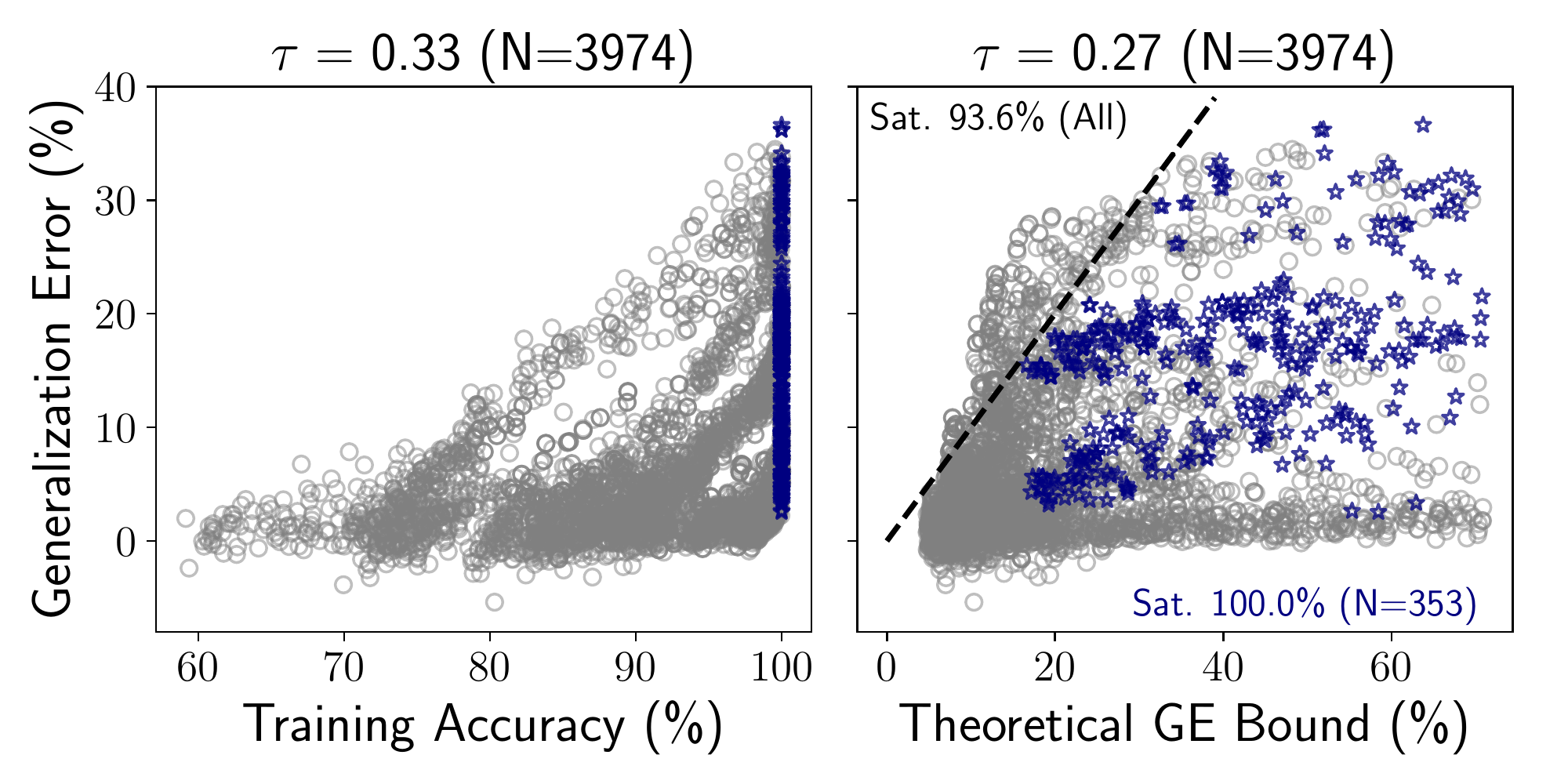}
\caption{}\label{fig:eurosat-v5}
\end{subfigure}
\begin{subfigure}{\textwidth}
\centering
\includegraphics[width=.8\textwidth]{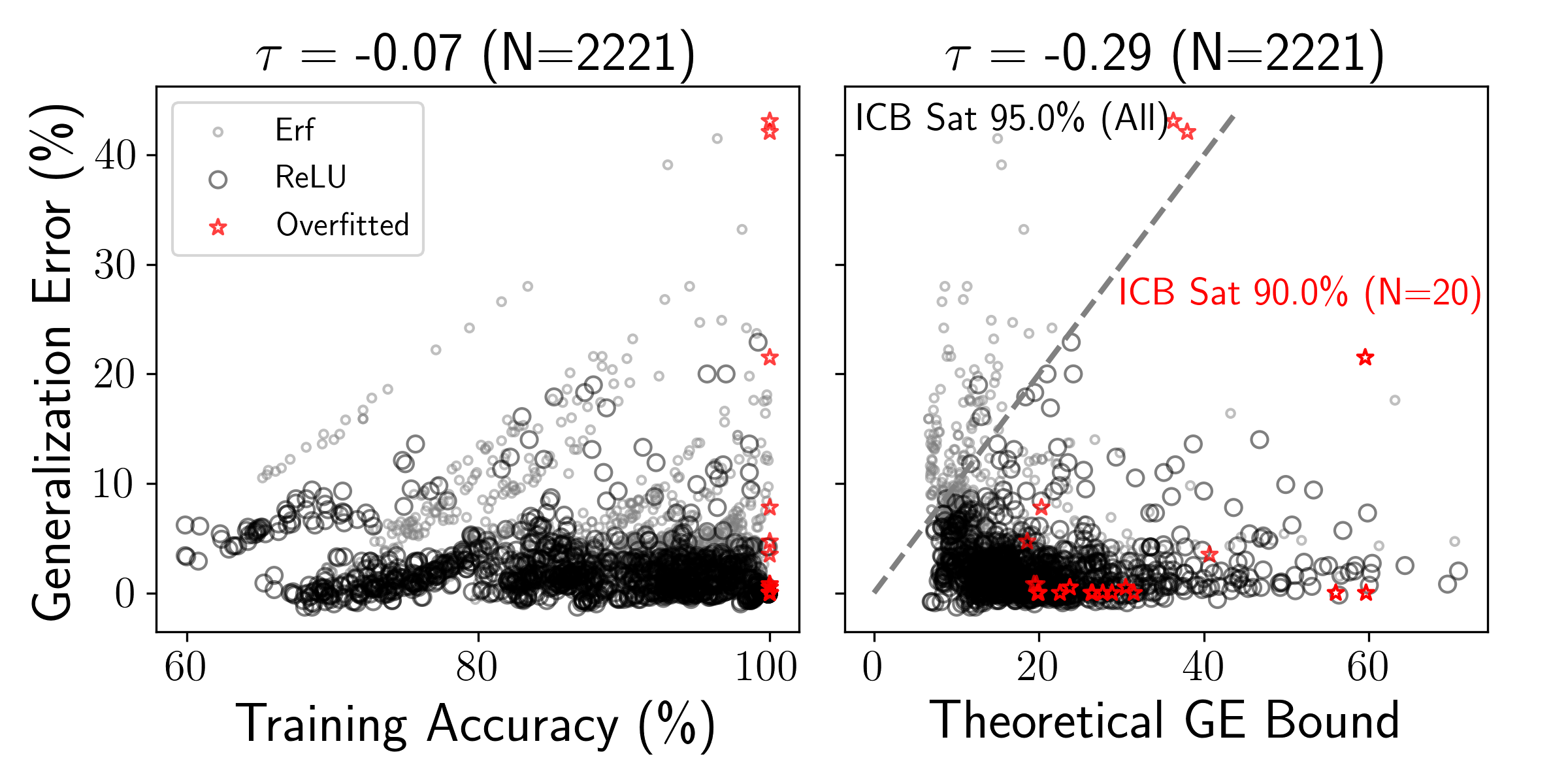}
\caption{}\label{fig:eurosat-v2-steady}
\end{subfigure}
\begin{subfigure}{\textwidth}
\centering
\includegraphics[width=.8\textwidth]{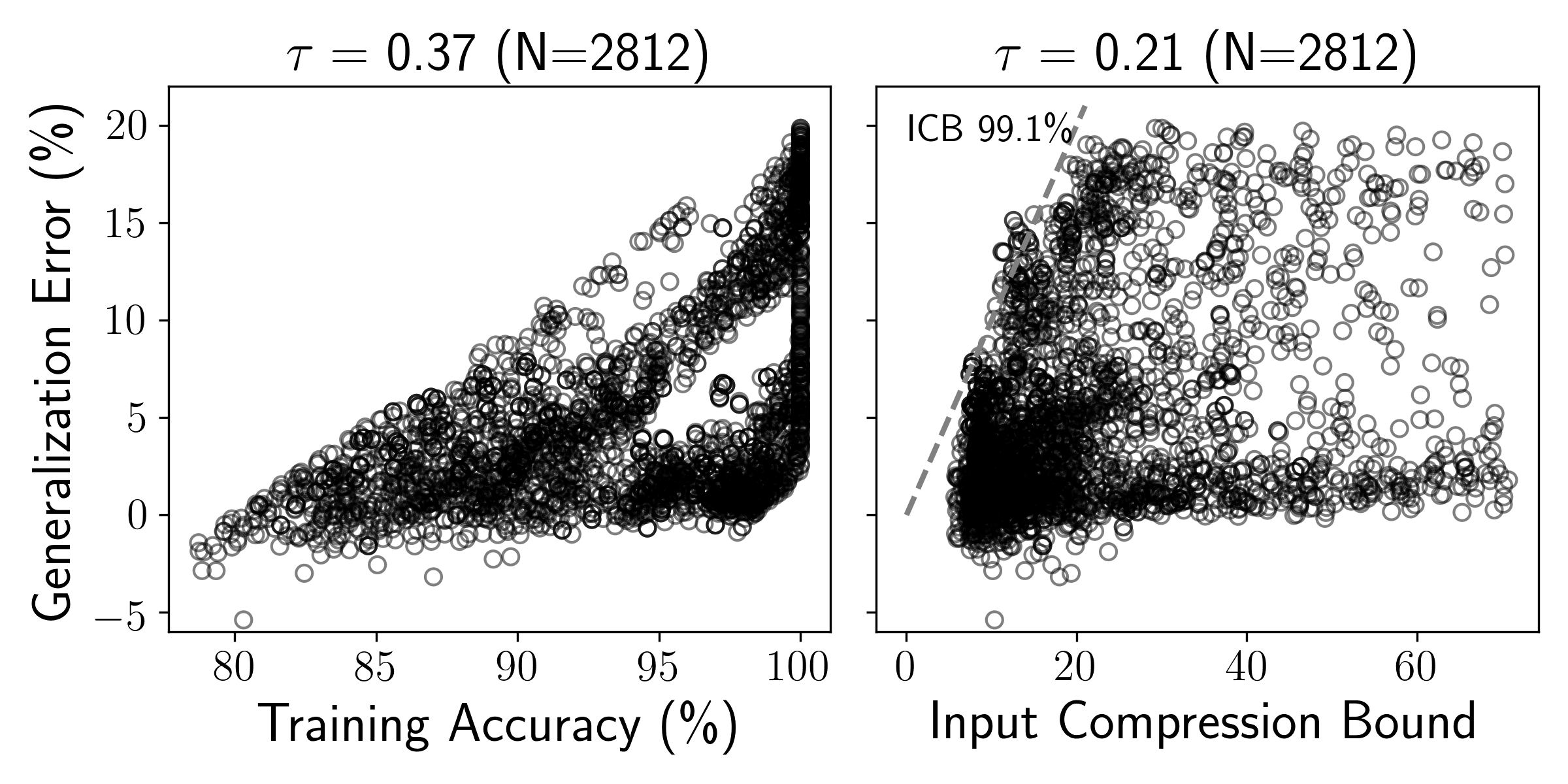}
\caption{}\label{fig:eurosat-v4-test-acc-80}
\end{subfigure}
\caption{\subref{fig:eurosat-v5}) EuroSAT model evaluated throughout training,
\subref{fig:eurosat-v2-steady}) models evaluated at steady state \(t=\infty\),
\subref{fig:eurosat-v4-test-acc-80}) select models from \subref{fig:eurosat-v5}) with
$\geq 80 \%$ test accuracy.}
\label{fig:eurosat-scatter}
\end{figure}

\begin{table}[]
\centering
\caption{Kendall's \(\tau\) ranking for~\texttt{EuroSAT}. See complete caption in~\S\ref{sec:sub-appendix-through-training}.}
\label{tab:eurosat-detailed}
\begin{tabular}{ccccccccc}
\toprule
\multicolumn{3}{c}{Details} & \multicolumn{3}{c}{Train (baseline)} & \multicolumn{3}{c}{ICB} \\ \midrule
Task & $N_\text{valid}$ & ICB \% & Clean & AWGN & FGSM & Clean & AWGN & FGSM \\ \midrule
0 vs. 1 & 414 & 100\% & 0.26 & 0.40 & 0.37 & 0.25 & 0.43 & 0.38 \\
1 vs. 2 & 389 & 100\% & 0.30 & 0.59 & 0.50 & 0.24 & 0.54 & 0.37 \\
2 vs. 3 & 468 & 68\% & 0.86 & 0.91 & 0.76 & 0.77 & 0.78 & 0.60 \\
3 vs. 4 & 490 & 97\% & 0.86 & 0.86 & 0.83 & 0.72 & 0.72 & 0.65 \\
4 vs. 5 & 485 & 100\% & 0.62 & 0.61 & 0.77 & 0.70 & 0.70 & 0.68 \\
5 vs. 6 & 444 & 100\% & 0.78 & 0.79 & 0.80 & 0.68 & 0.67 & 0.62 \\
6 vs. 7 & 475 & 72\% & 0.89 & 0.88 & 0.80 & 0.74 & 0.74 & 0.62 \\
7 vs. 8 & 467 & 100\% & 0.80 & 0.83 & 0.82 & 0.77 & 0.77 & 0.67 \\
8 vs. 9 & 335 & 100\% & 0.47 & 0.74 & 0.63 & 0.40 & 0.61 & 0.51 \\ \midrule
\multicolumn{2}{r}{Row average} & \multirow{2}{*}{93\%} & \textbf{0.65} & \textbf{0.73} & \textbf{0.70} & 0.59 & 0.66 & 0.57 \\
\multicolumn{2}{r}{Overall} &  & \textbf{0.34} & \textbf{0.36} & \textbf{0.33} & 0.26 & 0.28 & 0.28 \\ \bottomrule
\end{tabular}
\end{table}

\subsection{Bounding generalization at steady state}
\label{sec:sub-appendix-steady state}

Results for bounding the generalization error at steady state
are summarized in Table~A\ref{tab:ten-choose-two-bounding-error-steady-state}.

\begin{table}
\caption{\textbf{Results for bounding generalization error across five (5) datasets with~\acrfull{icb}.}
For each dataset, all ${\binom{10}{2}}=45$ binary label combinations are
evaluated for $S=25-50$ meta-parameter combinations drawn uniform random.
The \gls{icb} \% column indicates the percentage of $N_{\text{trials}}$ that satisfy the~\gls{icb},
where $N_{\text{trials}} = {\binom{10}{2}} \times S$.
The mean and maximum Clean generalization error is indicated
by ``Mean Err'' and ``Max Err'' respectively.
The training set sample size is indicated by $N_{\text{train}}$.
A test set with $N_{\text{test}}=2000$ is used in all cases except for~\texttt{EuroSAT} ($N_{\text{test}}=1000$).
Results are broken down by nonlinearity type as~\texttt{Erf} resulted
in~\gls{icb} being satisfied less often.}
\centering
\begin{tabular}{lrlrrrr} 
\toprule
& & & &  & \multicolumn{2}{c}{Error} \\ \midrule 
\textbf{Dataset} & $N_{\text{train}}$ & $N_{\text{trials}}$ & Arch & \gls{icb} \% & Mean Err & Max Err \\ \midrule 
\multirow{2}{*}{~\texttt{MNIST}} & \multirow{2}{*}{1000} & \multirow{2}{*}{1125} & \texttt{Erf}  & 100.0 & 0.8 & 2.4 \\ 
 & & & \texttt{ReLU} & 100.0 & 0.8 & 2.8 \\ \midrule 
\multirow{2}{*}{\texttt{Fashion}} & \multirow{2}{*}{1000} & \multirow{2}{*}{2250} & \texttt{Erf} & 99.9 & 1.3 & 16.5 \\ 
 & & & \texttt{ReLU}  & 100.0 & 1.4 & 17.7 \\ \midrule 
\multirow{2}{*}{\texttt{SVHN}} & \multirow{2}{*}{2000} & \multirow{2}{*}{2250} & \texttt{Erf}  & 0.3 & 14.1 & 35.7 \\ 
 & & & \texttt{ReLU} & 52.8 & 7.4 & 24.8 \\ \midrule 
\multirow{2}{*}{\texttt{CIFAR-10}} & \multirow{2}{*}{2000} & \multirow{2}{*}{2250} & \texttt{Erf} & 47.6 & 8.5 & 39.5 \\ 
 & & & \texttt{ReLU} & 86.5 & 5.4 & 36.3 \\ \midrule 
\multirow{2}{*}{\texttt{EuroSAT}} & \multirow{2}{*}{1000} & \multirow{2}{*}{1125} & \texttt{Erf} & 90.0 & 2.1 & 22.9 \\ 
 & & & \texttt{ReLU} & 99.5 & 5.1 & 43.0 \\ \bottomrule 
\end{tabular}
\label{tab:ten-choose-two-bounding-error-steady-state}
\end{table}

\begin{figure}
\centering
\includegraphics[width=\textwidth]{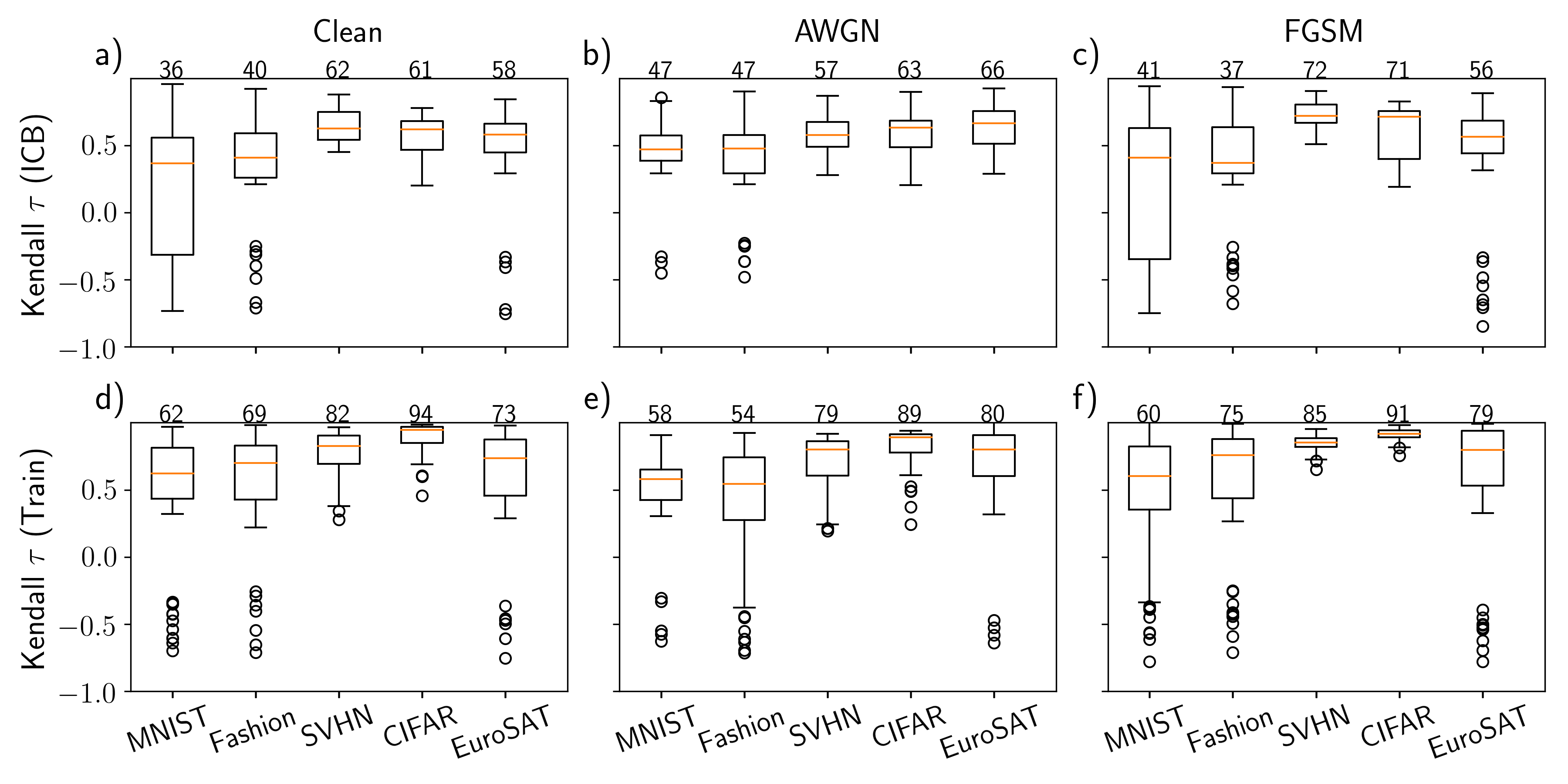}
\caption{Boxplots show the Kendall $\tau$ ranking between:
a,d) Clean, b,e) AWGN, and c,f) FGSM~\glspl{ge} and~\gls{icb} (\emph{top row})
compared to a training accuracy baseline (\emph{bottom row}).
We discard $\tau$ values with corresponding $p > 0.05$.
The median $\tau$ value is annotated above each box and multiplied by 100
for ease of interpretation.}
\label{fig:ranking_gen_err_steady_state}
\end{figure}

\subsection{Advantage of~\gls{icb} versus~\gls{mi}}
\label{sec:appendix-mi-vs-icb}

\begin{figure}
\centering
\begin{subfigure}{\textwidth}
\includegraphics[width=\textwidth]{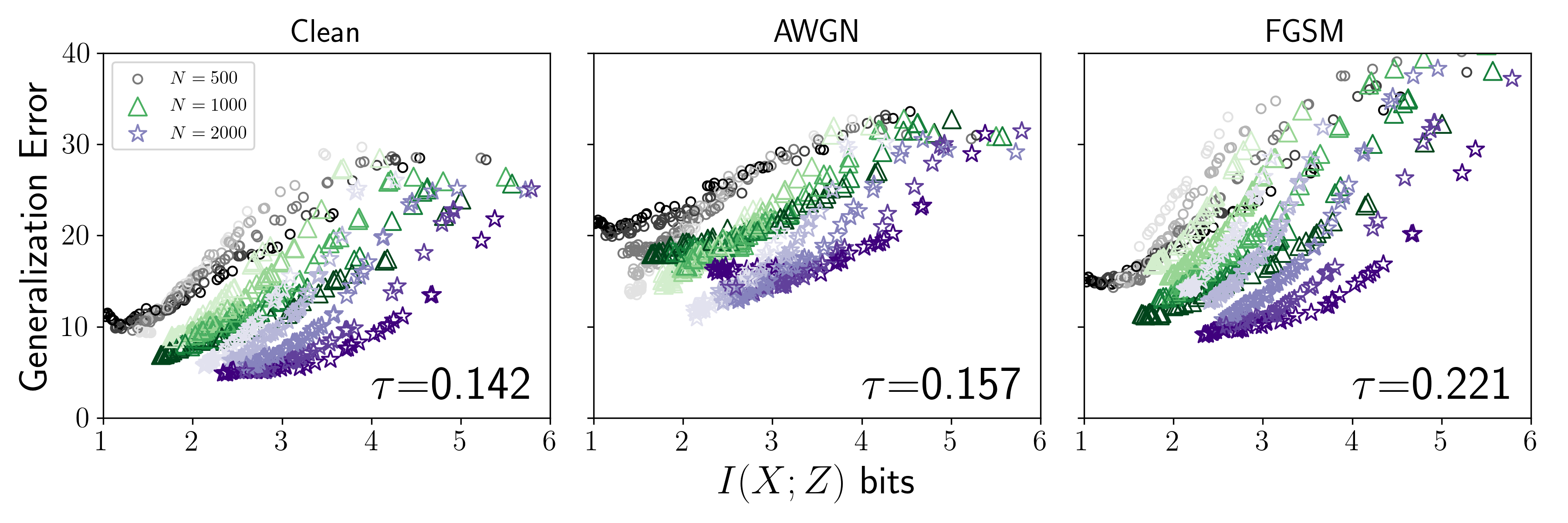}
\end{subfigure}
\begin{subfigure}{\textwidth}
\includegraphics[width=\textwidth]{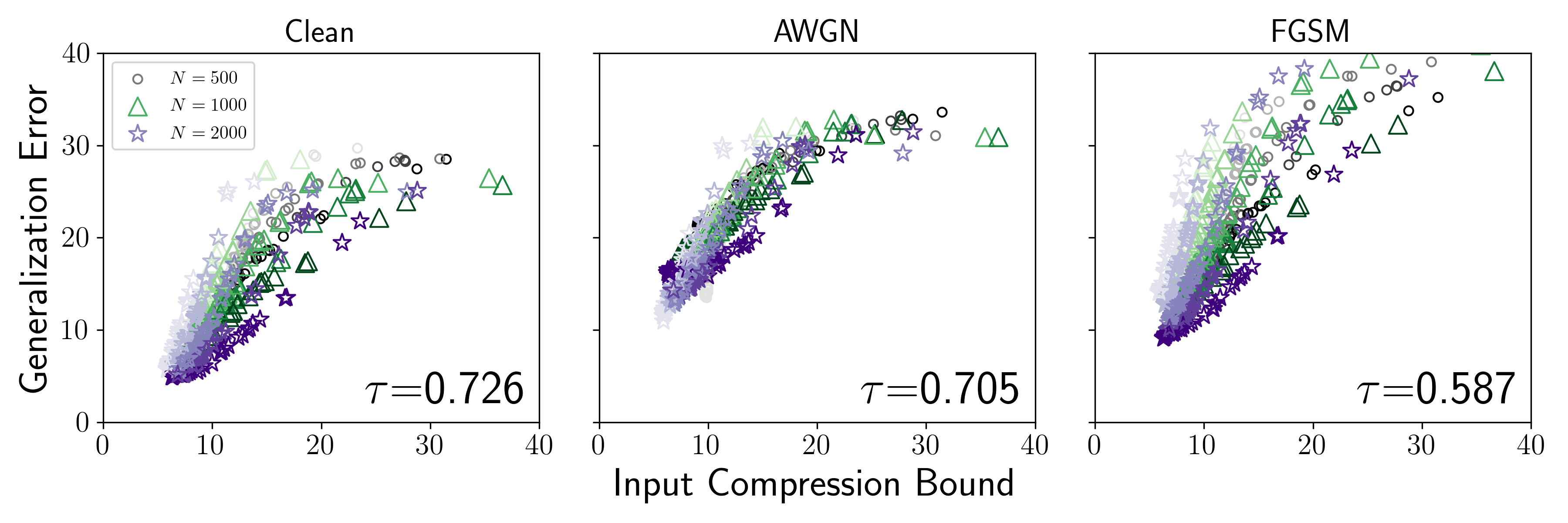}
\end{subfigure}
\caption{\textbf{\gls{icb} (bottom) ranks~\glspl{ge} better 
than $I(X; Z)$ alone (top) for different training set sizes.} 
Shown are 750 fully-connected \gls{ntk} ReLU models trained ($t=\infty$) 
on a~\texttt{CIFAR-10} binary classification task (classes 2 and 5) 
using three different training set sizes of $N=\{500, 1000, 2000\}$ 
and a test set with $N=2000$. 
For each training set, 250 meta-parameter combinations are drawn 
from a uniform random distribution (see~\S\ref{sec:sub-exp-steady} for 
details). 
Model depth is indicated by the colour intensity for each series, where the darkest 
shade indicates the maximum depth of five (5) layers. Three~\gls{ge} types are evaluated: 
\texttt{Clean} (standard), \gls{awgn} (adversarial), and \gls{fgsm} (adversarial) are plotted with 
respect to $I(X; Z)$ (\emph{top row}) and \gls{icb} (\emph{bottom row}). Plotting~\gls{ge} versus 
the \gls{icb} better aligns results for different sized training sets ($N$) compared to $I(X; Z)$, 
and yields a better ranking in terms of Kendall-$\tau$.}
\label{fig:mi-vs-icb}
\end{figure}

To gain further insight into~\gls{icb}, we examine~\glspl{ge} for 
a specific~\texttt{CIFAR-10} binary classification task (classes 2 and 5) 
using three different training set sizes. Plotting~\glspl{ge} 
with respect to \(I(X; Z)\) alone yields a poor overall ranking, 
whereas~\gls{icb} effectively aligns trials with different training set 
sizes~(Figure~\ref{fig:mi-vs-icb}).

\end{document}